\documentclass[12pt, a4paper]{article}
\usepackage[english, american]{babel}
\usepackage[utf8]{inputenc}     % latin1 (for windows-1252, ASCII), or utf8
\usepackage{amsmath}
\usepackage{amssymb}
\usepackage{graphicx, subcaption}
\graphicspath{{./Figures/}}
\usepackage{multirow}
\usepackage{booktabs}
\usepackage{xcolor}
\usepackage[linesnumbered,ruled,vlined] {algorithm2e}
\usepackage{url}
\usepackage{rotating}

\newcommand{\Eq}[1]{(\ref{#1})}
\newcommand{\Dict}{D}  % or \Delta ?

\newcommand{\lamvec}{\boldsymbol{\lambda}}
\newcommand{\phivec}{\boldsymbol{\phi}}
\newcommand{\UL}{U_i - \mathbf{u} \alpha \mathbf{v}^T}
\newcommand{\ULT}{U_i^T - \mathbf{v} \alpha \mathbf{u}^T}

\newcommand{\matlab}{{\textsc{Matlab}}}

\title{Kernel Recursive Least Squares Dictionary Learning Algorithm}

\author{Ghasem Alipoor$^1$\\
        \texttt{\small alipoor@hut.ac.ir}
\and
Karl Skretting$^2$\\
        \texttt{\small karl.skretting@uis.no}
}
\date{%
    $^1$\small Electrical Engineering Department, Hamedan University of Technology, Hamedan 6516913733, Iran\\%
    $^2$\small Department of Electrical and Computer Engineering, IDE, University of Stavanger, N-4036, Stavanger, Norway\\[2ex]%
}

\begin{document}

\vspace{5mm}

\maketitle

\begin{abstract}
  An online dictionary learning algorithm for kernel sparse representation is developed in the current paper. In this framework, the input signal nonlinearly mapped into the feature space is sparsely represented based on a virtual dictionary in the same space. At any instant, the dictionary is updated in two steps. In the first step, the input signal samples are sparsely represented in the feature space, using the dictionary that has been updated based on the previous data. In the second step, the dictionary is updated. In this paper, a novel recursive dictionary update algorithm is derived, based on the recursive least squares (RLS) approach. This algorithm gradually updates the dictionary, upon receiving one or a mini-batch of training samples. An efficient implementation of the algorithm is also formulated. Experimental results over four datasets in different fields show the superior performance of the proposed algorithm in comparison with its counterparts. In particular, the classification accuracy obtained by the dictionaries trained using the proposed algorithm gradually approaches that of the dictionaries trained in batch mode. Moreover, in spite of lower computational complexity, the proposed algorithm overdoes all existing online kernel dictionary learning algorithms.

  \vspace{5mm}
  \noindent \emph{\textbf{Keywords:}} \emph{Sparse Representation; Online Dictionary Leaning; Kernel Methods; Recursive Least Squares; Classification}
\end{abstract}

\section{Introduction}\label{SecInt}
Data factorization methods have met with considerable success
in discovering latent features of the signals encountered in wide-ranging applications.
In this way, the representation bases, which make up the columns of the basis matrix or dictionary,
are learned from the available samples of the target environment.
An example is the sparse representation (SR) in which the dictionary is intended to
best represent the data with a small number of atoms, much smaller than the dimension of the signal space.
It has been shown that, in addition to a more informative representation of signals,
imposing sparsity constraints on the representation coefficients can
improve the generalization performance and the computational efficiency \cite{BibHuang06, BibRubinstein10, BibZhang15a}.
Furthermore, the sparse representation is more robust to noise, redundancy, and missing data.
These features are mainly attributed to the fact that the intrinsic dimension of natural signals
is usually much smaller than their apparent dimension and hence
SR in an appropriate dictionary can extract these intrinsic features more efficiently.
SR has been a successful strategy and has received considerable attention and achieved state-of-the-art results
in many applications, e.g. vision, signal and image processing, and pattern recognition
\cite{BibHuang06, BibWright10, BibRubinstein10, BibZhang15a}.

SR entails two main problems. The first problem is sparse coding or Sparse Approximation (SA),
in which, a data sample is represented using a given dictionary, based on an optimization criterion.
Many optimization criteria and numerous solutions have been proposed for this purpose,
mostly with matching or basis pursuit algorithms \cite{BibMallat93, BibChen01, BibSkretting08}.
The second and more challenging problem is designing an effective and efficient dictionary \cite{BibRubinstein10}.
Dictionaries can be either pre-determined or data-driven.
It is well-accepted that data-driven dictionaries, learned from available training data samples
via a procedure called Dictionary Learning (DL), generally can better extract the inherent features
of signals and hence produce better results in many applications.
DL is generally a combinatorial and highly non-convex optimization problem with two outputs,
namely the learned dictionary and the representation coefficients.
Nevertheless, this problem is convex with respect to each of these two arguments
and is usually solved by alternating between two steps of
SA (fixing the dictionary) and dictionary update (fixing the representation coefficients).
Two main algorithms for DL are the Method of Optimal Directions (MOD) \cite{BibEngan99}
and the K-Singular Value Decomposition (K-SVD) \cite{BibAharon06}.

Classical SR-based methods represent data samples by linear combination of the atoms of the dictionary in the Euclidean space.
These linear models are incapable of coping with nonlinearity encountered in many real applications.
One intuitive remedy for this shortcoming is utilizing the kernel methods
that have shown considerable success, in various fields \cite{BibShawe04, BibHofmann08}.
Many Kernel SR (KSR) and KSR-based Classification (KSRC) algorithms have also been derived.
To this end, data samples are first nonlinearly mapped into a higher (or even
infinite) dimensional Reproducing Kernel Hilbert Space (RKHS), known as the feature space,
and are then sparsely represented based on a dictionary in the same space.
A few examples of the recent applications of the KSRC are medical applications \cite{BibMoradi19, BibTong21},
image quality assessment \cite{BibZhou20}, image classification \cite{BibZhang21}, computer vision \cite{BibFan19},
domain adaption \cite{BibLei21} and  fault diagnosis \cite{BibGao21}.

Several linear SA algorithms have been straightforwardly extended to their kernel counterparts \cite{BibLi11a, BibYin12, BibNguyen13}.
A large number of studies reported on the KSR have only considered this problem and have not touched the DL problem,
where the dictionary is represented by a subset of the transformed training exemplars \cite{BibZhang11, BibYuan12, BibYin12}.
The Kernel DL (KDL) problem has been first addressed in \cite{BibGao10} where in the update step,
an approximate solution of the stochastic gradient descent approach is used to optimize the dictionary, atom by atom.
The SA and DL problems in the spaces of symmetric positive definite (SPD) matrices were addressed in \cite{BibHarandi12, BibLi13, BibWu15},
using the fact that the SPD matrixes form a Riemannian manifold and then embedding this manifold into an RKHS.
As a result of using the gradient descent in the original input space,
the application of the abovementioned methods is restricted to specific kernels.

The milestone in KDL is the Kernel Method of Optimal Direction (KMOD) algorithm proposed in \cite{BibNguyen13},
in which Van Nguyen et al. have shown that the solution of the KDL,
formulated as a Least-Squares (LS) minimization problem, can be represented by a linear combination of the training data samples.
In particular, having $L$ training data samples $\{x_l \in \mathbb{R}^N, \quad l=1, 2, \ldots,L\}$
that make up the columns of the matrix $X \in \mathbb{R}^{N \times L}$ , the optimum dictionary has the form
\begin{equation}
  \label{eqoptD}
  D^*=\phi(X)B \quad \in \mathbb{R}^{F \times Q},
\end{equation}
where $Q$ is the number of dictionary atoms and $F$ is the dimension of the feature space or the length of each atom in that space.
$\phi$ is a non-linear mapping from $\mathbb{R}^N$ to the feature space $\mathcal{F}$
and the coefficient matrix $B \in \mathbb{R}^{L \times Q}$ is optimized based on kernelized versions of K-SVD and MOD algorithms.
Hence, the complex problem of learning a dictionary in the high (or even infinite) dimensional feature space
can be simplified to finding the optimum matrix $B$, while staying in the original space.
To the best of our knowledge, this simplification was adopted in all other similar studies
\cite{BibThiagarajan14, BibChen15, BibWu15, BibBrien16, BibAbrol16}.

Batch learning algorithms, in both linear and kernelized scenarios, incur a     significant burden on computational and storage;
as they generally require access to the entire batch of training data in each iteration.
This problem is more severe in the case of KSR, where, as a result of the representer theorem,
the dictionary is represented by all training data samples.
This issue was addressed in \cite{BibAbrol16}, in which a hierarchical DL approach was proposed that learns
the matrix $B$ in the coefficient space, independent of any computations involving the kernel matrix;
but the Kernel SA stage still depends on the entire kernel matrix.
Additionally, the Linearized Kernel Dictionary Learning (LKDL) method proposed in \cite{BibGolts16} does not solve this problem,
as the complexity and the required storage still quadratically depend on the number of input samples.
Moreover, both K-SVD and MOD algorithms, and their kernelized versions,
suffer from the dilemmas of matrix inversion, namely complexity, and singularity.

To address these essential shortcomings, many algorithms have been proposed for online DL, but in the Euclidean space \cite{BibMairal10, BibSkretting10, BibNaderahmadian16}.
Online algorithms can better cope with the dynamic change of the problem in non-stationary environments
and have appealing applications in real-time problems, such as prediction and tracking problems.
They have also been found useful in batch scenarios where the amount of the training data is too large to be accessed and processed at once.
Among the proposed online linear DL algorithms, the current study more relates to the RLS DL algorithm proposed in \cite{BibSkretting10}
which is essentially an iterative version of the MOD algorithm.
Nonetheless, a limited number of studies have been reported on the online KDL and this field of research is still infant.
Reference \cite{BibZhang15b} studies the problem of online dictionary learning on the SPD manifolds.
The proposed algorithm is an online version of the batch DL algorithm proposed in \cite{BibHarandi12} and hence is restricted to the Stein kernel.
A few online KDL algorithms have also been proposed in \cite{BibBrien16, BibLiu16, BibLee16} that they all rely on the representer theorem \eqref{eqoptD}.
Akin to other online kernel methods that use this theorem, it is necessary to duly control the size of the involved data samples;
the set that we refer to it along with all corresponding variables as the profile.

The sparse KDL (sKDL) algorithm was proposed in \cite{BibBrien16} in which a row-sparsity constraint is imposed
on the coefficient matrix $B$ to restrict the dictionary to be represented based on a few training samples.
This algorithm is then extended to online scenarios,
where the profile size is limited by selecting the incoming samples based on the reconstruction error.
A similar row-sparsity constraint idea is used in the Online KDL with Row-Sparsity (OKDLRS) algorithm developed in \cite{BibLiu16}, where a new model is proposed to simultaneously perform profile selection and kernel dictionary learning.
The formulated non-convex and non-smooth optimization problem is solved by proximal alternating linearized minimization.
Two slightly different Online KDL algorithms with a Fixed Budget (OKDLFB) are also proposed in \cite{BibLee16}
based on the Stochastic Gradient Descent (SGD) method.
The difference between these two algorithms is in the space, whether feature space or parameter space, in which the dictionary update is performed.

In the current study, an online kernel dictionary learning algorithm is developed
that recursively solves the LS minimization problem, with the aid of the Matrix Inversion Lemma (MIL).
This algorithm is an extension of the RLS DL algorithm proposed in \cite{BibSkretting10} to the feature spaces;
however with very different concepts and challenges.
This extension is done in a way similar to that of the work reported in \cite{BibEngel04} that kernelizes the RLS adaptive filtering algorithm for regression,
but in a clearly different context and with absolutely different derivations.
On the other hand, our algorithm can be viewed as an online extension to the batch KMOD algorithm of \cite{BibNguyen13}.
The derived algorithm is formulated efficiently and
various strategies are also examined for restricting the size of the profile.
The proposed algorithm differs from all previous online KDL algorithms in that,
instead of the gradient method, our algorithm is based on the RLS approach.
It is well-known that RLS algorithms are an order of magnitude faster in convergence than the gradient-based ones \cite{BibLiu10, BibHeykin14}.
Furthermore, our algorithm is not restricted to a specific kernel.

In summary, the main contributions of this study are as follows:
\begin{enumerate}
    \item A novel online KDL algorithm is developed that is the first one that is based on the RLS method.
    \item A relatively efficient implementation of the algorithm is derived.
    \item The performance of the proposed algorithm in classification is studied and compared
            with those of other KDL algorithms over four disparate practical datasets.
\end{enumerate}

After the introduction and the literature review presented in this section,
the remaining sections of the paper are organized as follows.
The original batch KDL problem is formulated in section \ref{SecKLS} and its solution is derived therein.
In section \ref{SecOnline}, this algorithm is extended to the online mode, by recursively updating the dictionary;
all relevant aspects are also covered in this section.
Experimental results are reported in section \ref{SecTest} and the paper is then concluded in the last section.

\vspace{5mm}
\noindent \textbf{Notation review}
\vspace{2mm}

In this paper, we assume the following sizes: $s < Q \le L \le L_{max} \le L_{tot}$ and $F \le \infty$.
$s$ is the number of non-zero elements in the SA of each data vector,
$Q$ is the number of atoms in the dictionary $\Dict$,
$L$ is the number of training vectors that are (still) remembered,
$L_{tot}$ is the total number of available training vectors.
$F$ is the dimensionality of the feature space that depends on the kernel used and can be rather large, e.g., for polynomial kernels, or even infinite, as for RBF kernels.
Furthermore, $N$ is the dimensionality of the feature space or the number of elements in each data vector,
and $M$ is the number of training input data samples in each mini-batch.
As we will see, $L$ can increase to $L_{max}$ as subsequent training vectors are processed.
To make the described algorithm more intelligible,
the dimensions of all the variables involved in the algorithm are summarized in Table \ref{TabNot}.
In this table, all variables are shown without any subscript $i$;
moreover, these variables may be used in the text with different subscripts or diacritical marks.

\begin{table} [h]
\centering
    \caption{Dimensions of the variables involved in deriving the algorithm (Variables that are not described in the third column are the intermediate variables defined throughout the derivations.)}
    \resizebox{\textwidth}{!}{
        \begin{tabular}{lll}
            Variable & Dimension & Description \\ \hline \hline
            $\textbf{x}$ & $N \times M$ & a mini-batch of $M$ input vectors in the original space \\
            $X$ & $N \times L$ & input matrix in the original space \\
            $\phivec$, \ $\textbf{r}$ & $F \times M$ & input and error vectors in the feature space \\
            $\Phi,$ \ $R$ & $F \times L$ & input and error matrixes in the feature space \\
            $D$ & $F \times Q$ & virtual dictionary in the feature space \\
            $\textbf{w}$, \ $\textbf{u}$, \ $\textbf{h}$ & $Q \times M$ & $\textbf{w}$ is (are) the coefficient vector(s). \\
            $W$, \ $U$ & $Q \times L$ & $W$ is the coefficient matrix. \\
            $C$, \ $\Psi$ & $Q \times Q$ & $C=(WW^T + \xi I )^{-1}$ and $\Psi$ is the dictionary Gram matrix. \\
            $\textbf{k}$, \ $\textbf{v}$ & $L \times M$ & $\textbf{k}$ is (are) the kernel vector(s). \\
            $\lamvec$ & $L \times 1$ & forgetting factor vector\\
            $K$, \ $\Lambda$ & $L \times L$ & kernel and diagonal forgetting factor matrixes \\
            $\sigma_\textbf{x}$, \ $\alpha$ & $M \times M$ & $\sigma_\textbf{x} = k(\textbf{x},\textbf{x})$\\
            $\lambda$ & $1 \times 1$ & forgetting factor
        \end{tabular}}
    \label{TabNot}
\end{table}

\section{Kernel Least Squares Dictionary Learning}\label{SecKLS}

The purpose of a dictionary learning algorithm for KSR is to find
an optimum dictionary for sparse representation of the input sample vectors, in a feature space
induced by a kernel function.
This problem can be solved iteratively, at each instant, in two steps.
\begin{itemize}
  \item In the first step, having the dictionary updated in the preceding instant, SA of the mapped input vector is found.
      For this end, we use the KORMP algorithm, which is a kernelized version of the Order Recursive Matching Pursuit (ORMP) algorithm proposed in \cite{BibSkretting08}.\\

  \item In the second step, having the coefficients for sparse representation of previous samples, the dictionary is updated.
      For this purpose, we develop a recursive algorithm based on the LS criterion.
\end{itemize}
The recursive solution to the LS problem is derived along the same lines as
the RLS and the Kernel RLS (KRLS) adaptive filtering algorithms \cite{BibEngel04}
and the RLS DL algorithm \cite{BibSkretting10}.
In this section, the objective function of the KDL is defined
and the direct solution to the optimum dictionary is derived.

\subsection{LS Solution}\label{SubSecLS}

At each instant, the objective is to find a (virtual) dictionary in the feature space
\begin{equation}
  \label{eqDict}
  \Dict = [\mathbf{d}_1, \mathbf{d}_2, \ldots, \mathbf{d}_Q] \quad \in \mathbb{R}^{F \times Q}
\end{equation}
for sparse representation of the input vectors up to the current time.
In particular, we aim at a virtual dictionary $D$ that the input vectors mapped into the feature space
can be represented by a linear combination of a small number of atoms of that dictionary.

At each time instant, the memory consists of $L$ input sample vectors
$\{\mathbf{x}_j\}_{j=1}^L$ which can be represented as columns of a matrix
\begin{equation}
  \label{eqX}
  X = [\mathbf{x}_1, \mathbf{x}_2, \ldots, \mathbf{x}_L]
  \quad \in \mathbb{R}^{N \times L}.
\end{equation}
The input vectors can, at least virtually, be mapped into the feature space
$\phivec_j = \varphi(\mathbf{x}_j)$ and these feature vectors
can be represented as columns of a matrix
\begin{equation}
  \label{eqPhi}
  \Phi = [\varphi(\mathbf{x}_1), \varphi(\mathbf{x}_2), \ldots, \varphi(\mathbf{x}_L)]
       = [\phivec_1, \phivec_2, \ldots, \phivec_L]
  \quad \in \mathbb{R}^{F \times L}.
\end{equation}

Moreover, we assume that, for each feature vector, the coefficients for a sparse representation
in the feature space, i.e. $\{\mathbf{w}_j\}_{j=1}^L$, are available.
The coefficient matrix is therefore defined as:
\begin{equation}
  \label{eqW}
  W = [\mathbf{w}_1, \mathbf{w}_2, \ldots, \mathbf{w}_L]
  \quad \in \mathbb{R}^{Q \times L}.
\end{equation}
These coefficients are typically found by SA, e.g. using the KORMP algorithm,
based on a dictionary from an earlier step/time in the DL process;
perhaps one that is close to the optimal dictionary $\Dict^*$.

Assuming that the approximation coefficients are available, we look only at the
dictionary update step in the iterative two-step DL procedure.
This problem can be described as the following regularized (non-weighted) LS minimization problem

\begin{eqnarray}
  \label{eqLS1}
  \Dict^* & = & \arg \min_{\Dict} \sum_{j=1}^L ||\mathbf{r}_j||_2^2 + \gamma ||\Dict||_F^2 \nonumber \\
        & = & \arg \min_{\Dict} \sum_{j=1}^L ||\phivec_j - \Dict \mathbf{w}_j||_2^2 + \gamma ||\Dict||_F^2.
\end{eqnarray}
$||\cdot||_F$ denotes the Frobenius norm, also known as the trace norm, and
$\mathbf{r}_j = \phivec_j - \Dict \mathbf{w}_j$ is the approximation error of the $j$th input sample, in the feature space, using the dictionary $D$.
Furthermore, $\gamma$ is a regularization parameter to trade-off between the representation error and the energy of the atoms.
Note that the first 2-norm is in the feature space.
The regularization effect of $\gamma$ is more easily seen from equation \Eq{eqLS}; adding a non-zero value is vital when $WW^T$ is nearly non-invertible.
Furthermore, a small non-zero $\gamma$ is helpful for making the normalization step (described in subsection \ref{SubSecNorm}) needed less frequently.

We may also define
\begin{equation}
  \label{eqR}
  R = [\mathbf{r}_1, \mathbf{r}_2, \ldots, \mathbf{r}_L]
  \quad \in \mathbb{R}^{F \times L}
\end{equation}
and express the LS problem \Eq{eqLS1} by matrixes in the Frobenius norm as:
\begin{eqnarray}
  \label{eqLS}
  \Dict^* & = & \arg \min_{\Dict} || R ||_{F}^2 + \gamma || \Dict ||_{F}^2 \nonumber \\
        & = & \arg \min_{\Dict} || \Phi - \Dict W ||_{F}^2 + \gamma || \Dict ||_{F}^2.
\end{eqnarray}
The solution of \Eq{eqLS} is
\begin{equation}
  \label{eqLSsol1}
  \Dict^* = (\Phi W^T) (W W^T + \gamma I )^{-1} = (\Phi W^T) C,
\end{equation}
where $C = (W W^T + \gamma I )^{-1}$ and $I$ is the identity matrix. Alternatively
\begin{equation}
  \label{eqLSsol2}
  \Dict^* = \Phi U^T,
\end{equation}
where $U = C W = (W W^T + \gamma I)^{-1} W \quad \in \mathbb{R}^{Q \times L}$. We see that each dictionary atom is a linear combination of the columns of $\Phi$ defined in \Eq{eqPhi},
i.e. the data vectors mapped into the feature space.
This is in accordance with model \eqref{eqoptD} which was first proposed in \cite{BibNguyen13};
in the current work, $B=U^T \in \mathbb{R}^{L \times Q}$ is found adaptively.

\subsection{Weighted LS Solution}\label{SubSecWLS}
Each term in \Eq{eqLS1} can be weighted with a given weight $\lambda_j$.
Adjusting $\lambda_j$ appropriately can control the memory span of the algorithm,
make the algorithm less dependent on the initial dictionary, and
improve both convergence properties as well as the representation ability of the resulting dictionary.
All these weights can be collected into a column vector $\lamvec$ of length $L$ or along the diagonal in a
square diagonal matrix $\Lambda = \operatorname{diag}(\lamvec) \ \in \ \mathbb{R}^{L \times L}$.

\begin{equation}
  \label{eqlambda}
  \lamvec = [\lambda_1, \lambda_2, \ldots, \lambda_L]^T
  \quad \in \mathbb{R}^{L \times 1},
\end{equation}
\begin{equation}
  \label{eqLambda}
  \Lambda = \operatorname{diag}(\lamvec) \quad \in \mathbb{R}^{L \times L}.
\end{equation}

The regularized weighted LS (WLS) problem is defined as:
\begin{equation}
  \label{eqWLS1}
  \Dict^* = \arg \min_{\Dict} || (\Phi - \Dict W) \, \Lambda^{0.5} \, ||_F^2 + \xi ||D||_{F}^2,
\end{equation}
where $\Lambda^{0.5}$ is a matrix whose elements are square roots of the elements of $\Lambda$.
For now, $\xi$ can be set to $\gamma$. It is noted that
\begin{equation}
  \label{eqWLS2}
    || (\Phi - \Dict W) \, \Lambda^{0.5} \, ||_F^2 = \sum_{j=1}^L \left \{ \lambda_j ||(\phivec_j - \Dict \mathbf{w}_j)||_2^2 \right \}.
\end{equation}

The WLS solution is
\begin{eqnarray}
  \label{eqD}
  \Dict = (\Phi \Lambda W^T) \left(W \Lambda W^T + \xi I \right)^{-1} = (\Phi \Lambda W^T) C = \Phi U^T,
\end{eqnarray}
where $C$ and $U$ matrixes are defined as:
\begin{equation}
  \label{eqC}
  C = \left(W \Lambda W^T + \xi I\right)^{-1}  \quad \in \mathbb{R}^{Q \times Q},
\end{equation}
\begin{equation}
  \label{eqU}
  U = C W \Lambda = \left(W \Lambda W^T + \xi I \right)^{-1} W \Lambda \quad \in \mathbb{R}^{Q \times L}.
\end{equation}
Finally, the reconstruction in the feature space is
\begin{equation}
  \label{eqPhiHat}
  \tilde{\Phi} = \Dict W = \Phi U^T W = \Phi \, \Lambda W^T \left(W \Lambda W^T + \xi I \right) ^{-1} W.
\end{equation}
Note that it is not generally possible to go from $\tilde{\Phi}$ to $\tilde{X}$, i.e. the reconstruction in the original input space.
However, this representation is still meaningful and sensible;
at least for applications in which signal reconstruction is not required,
e.g. feature extraction and classification applications.

\subsection{Kernel Sparse Approximation}\label{SubSecKSA}
One can perform Kernel sparse approximation without forming the dictionary in the feature space.
For instance, the ORMP algorithm, derived in \cite{BibSkretting08}, can be extended to the kernel space straightforwardly.
To this end, the input variables of the ORMP algorithm in the signal domain,
including the dictionary $D$ ($F$ in \cite{BibSkretting08}) and the signal vector $\mathbf{x}$,
should be replaced with the corresponding variables in the feature space.
More particularly, the kernelized ORMP algorithm, denoted as the KORMP algorithm, uses the dictionary Gram matrix $\Psi$,
the inner product of the dictionary and data vectors that is indicated by $\mathbf{h}$,
and $\sigma_x^2 = k(\mathbf{x},\mathbf{x})$ as input variables to solve, in the feature space,
for the approximation coefficient vector $\mathbf{w}$.

The dictionary Gram matrix $\Psi$ can be expressed as:
\begin{equation}
  \label{eqPsi1}
  \Psi = \Dict^T \Dict = U \Phi^T \Phi U^T = U K U^T
  \quad \in \mathbb{R}^{Q \times Q},
\end{equation}
where $K = \Phi^T \Phi \ \in \mathbb{R}^{L \times L}$
is a symmetric positive definite kernel matrix with elements
$K_{ij} = K_{ji} = \phivec_i^T \phivec_j = <\phivec_i,\phivec_j> = k(\mathbf{x_i},\mathbf{x_j})$.
Usually, the simple kernel function $k(\cdot)$ is given rather than the related,
and possibly more complicated, mapping function $\varphi(\cdot)$.
Let the new data vector be $\mathbf{x}$, without subscript, to distinguish
it from the data vectors in memory as stored in $X$.
The inner product vector $\mathbf{h}$ can be represented as:
\begin{equation}
  \label{eqh}
  \mathbf{h} = \Dict^T \varphi(\mathbf{x})
  = U \, \Phi^T \varphi(\mathbf{x}) = U \, \Phi^T \phivec = U \, \mathbf{k}
  \quad \in \mathbb{R}^{Q \times 1},
\end{equation}
where $\mathbf{k} = \Phi^T \phivec \ \in \mathbb{R}^{L \times 1}$,
and $k_j = \phivec_j^T \phivec = \ <\phivec_j,\phivec> \ = k(\mathbf{x_j},\mathbf{x})$.
Moreover,
\begin{equation}
  \label{eqsigmax2}
  \sigma_x^2 = \phivec^T \phivec = \ <\phivec,\phivec> \ = k(\mathbf{x},\mathbf{x}).
\end{equation}

Note that $\Psi$ is independent of $\mathbf{x}$, while $\mathbf{h}$ and $\sigma_x^2$
depend on $\mathbf{x}$. Using these variables, the KORMP algorithm returns
the sparse coefficients $\mathbf{w}$, with $s$ non-zero elements,
and also the squared error, in the feature space,
$\mathbf{r}^T \mathbf{r} = ||\mathbf{r}||_2^2$.

\section{Online Kernel Dictionary Learning}\label{SecOnline}
The common practice in online dictionary learning algorithms, such as the KRLS DL algorithm proposed here,
is to repeat a process of SA followed by an update of the dictionary,
for each new training vector $\mathbf{x}$ (or a mini-batch of training vectors) that is presented to the algorithm.
On the other hand, in kernel methods, the size of the involved matrixes,
such as $U$, $W$, and $K$ matrixes in the KRLS DL algorithm, grow unboundedly with time.
This makes it necessary to restrict the size of data by storing only the most informative samples
and even removing some previously-stored samples from the memory.
Therefore, the dictionary update step, performed by updating the involved matrixes or the profile, may require:
\begin{itemize}
  \item adding one new training vector or a (mini-) batch of new training vectors
  \item removing one or a (mini-) batch of the previously stored training vectors
  \item dictionary normalization, so that all atoms have unit $\ell_2-$norm.
\end{itemize}

The iteration always starts by SA of new input data samples which returns the new coefficient vector(s) $\mathbf{w}$.
The matrixes at the start are denoted by subscript $(\cdot)_i$, resulting after the previous iteration $i$.
The step we consider is iteration $i+1$ and the resulting matrixes are thus denoted by a subscript $(\cdot)_{i+1}$.
Intermediate matrixes may be denoted by the hat symbol $(\hat{\cdot})$.
For simplicity vectors for step $i+1$ are usually denoted without a subscript,
as $\mathbf{x}$ and $\mathbf{w}$.
The involved matrixes are updated depending on what action we want to do.
In the following sections, the effect of each action on the involved matrixes is described.

\subsection{Adding data vectors}\label{SubSecAdd}
One of the main steps in KRLS DL algorithm is to add the new training data vector and
perhaps also apply a forgetting factor on previous data.
Using a forgetting factor $\lambda \le 1$, the weights of the
previous approximation errors are reduced by this factor,
while the new approximation is added with weight 1.
Having $\lambda = 1$ is the dictionary update step without the forgetting factor.
Note that this step increases the size of some matrixes used in the KRLS DL algorithm.
Specifically, one term is added to the summation \Eq{eqWLS2} and $L$ is increased by one and also

\begin{equation}\label{eqlam_i1}
  \lamvec_{i+1} = [\ \lambda \lamvec_i^T, \ 1 \ ]^T
\end{equation}
and
\begin{equation}\label{eqXW_i1}
  X_{i+1} = [X_i, \mathbf{x}] \quad \textrm{and} \quad W_{i+1} = [W_i, \mathbf{w}].
\end{equation}
Another option is to add a mini-batch, a set of some few $M$ data vectors, instead of a single vector.
This can be done by letting $\mathbf{x}$ and $\mathbf{w}$ in \Eq{eqXW_i1}
be matrices of size $N \times M$ and $Q \times M$, respectively, and replacing the scalar element $1$
in \Eq{eqlam_i1} with the $M$-length row vector $I_M^T$.

Referring to \Eq{eqC}, to make it possible to derive a simple recursive update equation for $C_i$,
here $\xi_i$ is defined as:
\begin{equation}\label{eqXi_i}
  \xi_i = \gamma \prod_{\lambda_j \in \Omega_i} \lambda_j,
\end{equation}
where $\Omega_i$ is the set of forgetting factors that correspond to all input data samples that have engaged in dictionary learning, up to time instant $i$.
Note that when no input sample is pruned (as described in subsection \ref{SubSecRmv}),
$\prod_{\lambda_j \in \Omega_i} \lambda_j = \lamvec_i(1)$,
i.e. the first element in the $\lamvec_{i}$ vector which equals the product of all previously used $\lambda$ values.
With this definition, $C_{i}$ can be recursively updated as:
\begin{equation} \label{eqCi1}
  C_{i+1}^{-1} = \lambda \, C_i^{-1} + \mathbf{w}\mathbf{w}^T.
\end{equation}
The term $\mathbf{w}\mathbf{w}^T$ is a $Q \times Q$ matrix; when $M \le Q$, its rank can be up to $M$.

Applying Matrix Inversion Lemma (MIL):
\begin{equation}
\label{eqMIL}
  (A+EFG)^{-1} = A^{-1} - A^{-1} E (F^{-1} + G A^{-1} E)^{-1} G A^{-1}
\end{equation}
on \Eq{eqCi1} with $A=\lambda \, C_i^{-1}$, $E=\mathbf{w}$, $F=I_M$, and $G=\mathbf{w}^T$ gives
\begin{equation}
  C_{i+1} = \lambda^{-1} C_i \, - \, \lambda^{-1} C_i \mathbf{w}
            \big(I_M + \mathbf{w}^T \lambda^{-1} C_i \mathbf{w} \big)^{-1} \,
            \mathbf{w}^T \lambda^{-1} C_i. \nonumber
\end{equation}
As $(\lambda A)^{-1} = \lambda^{-1} A^{-1}$, this gives
\begin{equation}
  C_{i+1} = \lambda^{-1} \big( C_i \, - \, C_i \mathbf{w}
            (\lambda I_M + \mathbf{w}^T C_i \mathbf{w} )^{-1} \,
            \mathbf{w}^T C_i \big) \nonumber
\end{equation}

and

\begin{equation}
  \label{eqC_i1}
   C_{i+1} = \lambda^{-1} \, \big( C_i \, - \, \mathbf{u} \, \alpha \, \mathbf{u}^T \big),
\end{equation}
where,
\begin{equation}
  \label{eqalpha}
  \mathbf{u} = C_i \mathbf{w} \ \in \mathbb{R}^{Q \times M} \quad \textrm{and} \quad
  \alpha = \big( \lambda I_M + \mathbf{w}^T \mathbf{u} \big)^{-1} \ \in \mathbb{R}^{M \times M}.
\end{equation}
As $C_i$ is symmetric positive semi-definite (PSD),
$\mathbf{w}^T \mathbf{u} = \mathbf{u}^T \mathbf{w} = \mathbf{w}^T C_i \mathbf{w}$ and also $\alpha$
are $M \times M$ symmetric PSD matrixes.
Hence, assuming that $\mathbf{w}$ has full rank, these two matrixes are also full-rank.

Other relevant matrixes can also be recursively updated.
The symmetric (and PSD) kernel matrix $K$,
contains the results of evaluating the kernel function
over all pairs of training data vectors in matrix $X$.
By adding new data vectors, this matrix grows and its size increases
from $K_i$ of size $L_i \times L_i$ to
$K_{i+1}$ of size $L_{i+1} \times L_{i+1}$, where $L_{i+1} = L_{i} + M$.
To add the effect of the new input vector(s), matrix $K$ can be updated as:
\begin{equation}\label{eqK_i1}
  K_{i+1} = [\Phi_i, \phivec]^T [\Phi_i, \phivec]
        = \left[ \begin{array}{cc}   \Phi_i^T \Phi_i & \Phi_i^T \phivec \\
                  \phivec^T \Phi_i & \phivec^T \phivec   \end{array} \right]
        = \left[ \begin{array}{cc}   K_i & \mathbf{k} \\
                  \mathbf{k}^T & \sigma_{\mathbf{x}}^2   \end{array} \right],
\end{equation}
where $\phivec = \varphi(\mathbf{x}) \ \in \mathbb{R}^{F \times M}$,
$\mathbf{k} = \Phi_i^T \phivec \ \in \mathbb{R}^{L_i \times M}$, and
$\sigma_{\mathbf{x}}^2 = \phivec^T \phivec = k(\mathbf{x},\mathbf{x}) \ \in \mathbb{R}^{M \times M}$.

Matrix $U = C W \Lambda \ \in \mathbb{R}^{Q \times L}$, that is defined in \Eq{eqU},
also grows as $W$ grows and $L$ increases from $L_i$ to $L_{i+1} = L_{i} + M$.
Matrix $U$ is used to find vector $\mathbf{h}$ in \Eq{eqh} for SA and can be updated as (refer to appendix \ref{appendix:gU}):
\begin{equation} \label{eqUi1}
  U_{i+1} = \Big[ \, U_i - \mathbf{u} \alpha \, \mathbf{v}^T,
                           \, \mathbf{u} \alpha \, \Big].
\end{equation}
Note that $\lambda$ is scalar, $\alpha \in \mathbb{R}^{M \times M}$, and we have used $\alpha \mathbf{u}^T \mathbf{w} = I_M - \lambda \alpha$ from \Eq{eqalpha} and defined:
\begin{equation}
  \label{eqv}
  \mathbf{v}^T = \mathbf{u}^T W_i \Lambda_i, \quad
   \mathbf{v} = \Lambda_i W_i^T \mathbf{u}
   \quad \in \mathbb{R}^{L \times M}.
\end{equation}

If $\varphi(\cdot)$ function is accessible,
it is possible to update the dictionary from equation \Eq{eqD} as (refer to appendix \ref{appendix:gD}):
\begin{equation} \label{eqDi1}
  \Dict_{i+1} = \Dict_i + \mathbf{r} \alpha \mathbf{u}^T,
\end{equation}
since $\Phi_i \mathbf{v} = \Phi_i \Lambda_i W_i^T C_i \mathbf{w} = \Phi_i U_i^T \mathbf{w}
= \Dict_i \mathbf{w} = \tilde{\phivec}$ and $\mathbf{r} = \phivec - \tilde{\phivec}$.

Updating $\Psi$, which is PSD of size $Q \times Q$, is more complicated; referring to appendix \ref{appendix:gPsi}, it can be done as:
\begin{equation}  \label{eqPsi2}
  \Psi_{i+1} = \Psi_i + \mathbf{u} \alpha \mathbf{\tilde{u}}^T
             + \mathbf{\tilde{u}} \alpha \mathbf{u}^T
             + \mathbf{u} \alpha (\mathbf{v}^T K_i \mathbf{v}
               - \mathbf{v}^T \mathbf{k} - \mathbf{k}^T \mathbf{v}
               + \sigma_{\mathbf{x}}^2 ) \alpha \mathbf{u}^T,
\end{equation}
where we have defined:
\begin{equation}
   \label{equtilde}
   \mathbf{\tilde{u}} = U_i(\mathbf{k}-K_i\mathbf{v})
   \quad \in \mathbb{R}^{Q \times M}.
\end{equation}

Note that, for $M=1$, all these update equations involve only matrix-by-vector multiplications.
In this case, by recasting the initial matrix-by-matrix multiplication to only matrix-by-vector multiplications,
the complexity is reduced from $O(QL^2)$ to $O(L^2)$, since the largest matrix involved is the $L \times L$ matrix $K_i$.

\subsection{Removing data vectors}\label{SubSecRmv}
It may be required to remove one or a mini-batch of data sample(s), say $M$ vectors,  from the profile;
e.g. to make sure that the size of the profile does not exceed a predefined bound.
These data vectors should be selected properly.
The approach we use here to update the profile is to first remove the effect of these vectors in matrixes $C$, $D$, $U$, and $\Psi$.
Then we may choose, if we want, to zero out the corresponding $M$ columns in $X$, $W$, and $\Phi$;
and the corresponding $M$ columns and rows in matrixes $K$ and $\Lambda$, or to remove them.\\

We assume that the indices of $M$ data samples that are selected to be removed are collected in vector $\mathbf{m}$.
Defining $\lamvec_m = \lamvec_i(\mathbf{m})$, we start by setting the corresponding $M$ elements in $\hat{\lamvec}$ to 0, i.e.
$\hat{\lamvec}(\mathbf{m}) = 0$, and also $\hat{\Lambda}(\mathbf{m},\mathbf{m}) = \mathbf{0}$;
the other elements of $\hat{\lamvec}$ and $\hat{\Lambda}$ are as in $\lamvec_i$ and $\Lambda_i$, respectively.
Note that $\lamvec_m \in \mathbb{R}^M$ and $\hat{\Lambda}(\mathbf{m},\mathbf{m}) \in \mathbb{R}^{M \times M}$.
At first, nothing is done with $X_i$ and $W_i$ (and $\Phi_i$), the effect of
columns $\mathbf{m}$ is canceled by this change in $\hat{\Lambda}$.
Referring to \Eq{eqC}, setting $\hat{\lamvec}_m$ to zero thus gives
\begin{equation} \label{eqC_h}
  \hat{C}^{-1} = C_i^{-1} - \mathbf{w}_m \lamvec_m \mathbf{w}_m^T,
\end{equation}
where $\mathbf{w}_m = W_i(:, \mathbf{m}) \in \mathbb{R}^{Q \times M}$ contains the $M$ coefficient vectors that should be removed.
The special definition for the regularization factor \Eq{eqXi_i}, makes it possible to derive this recursive update equation.

Assuming that $\hat{C}^{-1}$ is (still) an invertible matrix,
using MIL \Eq{eqMIL} with $A=C_i^{-1}$, $F=-I_M$ and $E=G^T = \mathbf{w}_m \lamvec_m^{0.5}$ results in (refer to appendix \ref{appendix:pC}):
\begin{equation} \label{eqChat}
  \hat{C} = C_i + \mathbf{u}_m \alpha_m \mathbf{u}_m^T,
\end{equation}
in which
\begin{equation}
  \label{eqalpham}
  \mathbf{u}_m = C_i \mathbf{w}_m \quad \textrm{and} \quad
  \alpha_m = \big(\lamvec_m^{-1} - \mathbf{w}_m^T \mathbf{u}_m \big)^{-1}.
\end{equation}\\

The new $U$ matrix will be
\begin{equation}
  \label{eqUh}
  \hat{U} = \hat{C} W_i \hat{\Lambda}
   = C_i W_i \hat{\Lambda} + \mathbf{u}_m \alpha_m \mathbf{u}_m^T W_i \hat{\Lambda}
   = C_i W_i \hat{\Lambda} + \mathbf{u}_m \alpha_m \mathbf{v}_m^T,
\end{equation}
where $\mathbf{v}_m^T = \mathbf{u}_m^T W_i \hat{\Lambda}$
or if we prefer $\mathbf{v}_m = \hat{\Lambda} W_i^T \mathbf{u}_m$.
Note that $\mathbf{v}_m$ is defined similar to equation \Eq{eqv}, but elements $\mathbf{m}$ are set to zero
and $C_i W_i \hat{\Lambda}$ is defined similar to $U_i$ in \Eq{eqU}, with columns $\mathbf{m}$ equal to zero;
thus $\hat{U}$ equals $U_i + \mathbf{u}_m \alpha_m \mathbf{v}_m^T$, with columns $\mathbf{m}$ set to zero.
Equation \Eq{eqUh} can be rewritten as
\begin{equation}
  \label{eqUh2}
  \hat{U} = U_i - U_{\downarrow m} + \mathbf{u}_m \alpha_m \mathbf{v}_m^T,
\end{equation}
where $U_{\downarrow m} \ \in \mathbb{R}^{Q \times L}$ is a matrix for which $M$ columns whose indices are specified by $\mathbf{m}$ are the corresponding columns on $U_i$ and all other elements are zero.
Note that columns $\mathbf{m}$ in $\hat{U}$ are all-zero.

With this notation (refer to appendix \ref{appendix:pD1}),
\begin{equation} \label{eqDh1}
  \hat{D} = D_i - [\mathbf{\phi}_m \lamvec_m  - \Phi_i \mathbf{v}_m \alpha_m ]\mathbf{u}_m^T.
\end{equation}
Since $\hat{\Lambda} = \Lambda_i - \Lambda_{\downarrow m}$,
where  $\Lambda_{\downarrow m}$ is defined similar to $U_{\downarrow m}$ (refer to appendix \ref{appendix:phiv}),
\begin{equation} \label{Pshi_v}
  \Phi_i \mathbf{v}_m = \hat{\phivec}_m - \phivec_m + \phivec_m \lamvec_m \alpha_m^{-1}.
\end{equation}
We have used the fact that $\lamvec_m \mathbf{u}_m^T \mathbf{w}_m = I_M - \lamvec_m \alpha_m^{-1}$.
Inserting $\Phi_i \mathbf{v}_m$ from \Eq{Pshi_v} into \Eq{eqDh1} yields to (refer to appendix \ref{appendix:pD2}):
\begin{equation} \label{eqDh2}
  \hat{D} = D_i - \mathbf{r}_m \alpha_m \mathbf{u}_m^T.
\end{equation}
One can compare \Eq{eqDh2} with \Eq{eqDi1}.

The new $\Psi$ can also be updated as (refer to appendix \ref{appendix:pPhi}):
\begin{eqnarray} \label{eqPhi_h}
\hat{\Psi} & = & \Psi_i - (\mathbf{u}_m \hat{\mathbf{u}}_m^T + \hat{\mathbf{u}}_m \mathbf{u}_m^T) \nonumber \\
           &   & + \, \mathbf{u}_m [\lamvec_m \sigma_m^2 \lamvec_m - \lamvec_m \mathbf{k}_m^T \mathbf{v}_m \alpha_m - \alpha_m \mathbf{v}_m^T \mathbf{k}_m \lamvec_m \nonumber \\
           &   & + \, \alpha_m \mathbf{v}_m^T K_i \mathbf{v}_m \alpha_m] \mathbf{u}_m^T,
\end{eqnarray}
where $\hat{\mathbf{u}}_m \triangleq U_i [\mathbf{k}_m \lamvec_m - K_i \mathbf{v}_m \alpha_m] \in  \mathbb{R}^{Q \times M}$.
The same result can be achieved using \Eq{eqDh2} and the fact that $\hat{\Psi}_i = \hat{D}^T \hat{D}$.

Finally, the size of the involved matrixes $X_i$, $W_i$, $\Lambda_i$, $\Phi_i$ and $K_i$
can be reduced by discarding the columns $\mathbf{m}$, or the rows $\mathbf{m}$.
This prevents the size of these matrixes from growing unboundedly.
Again, for special case of $M = 1$, update equations involve only matrix-by-vector multiplications
and the complexity is of order $O(L^2)$.

\subsection{Profile abstraction} \label{SubSecAbs}

In the context of KSR, there are two complementary concepts of dictionary sparsification.
In kernel methods, the size of the network over which the signal is expanded,
or the number of past samples, increases with the size of the data.
This phenomenon is also clear from the algorithm derived in section \ref{SecKLS}, where the
size of the matrixes $X$, $W$, $K$, and $U$ grow linearly with time.
In other words, at each time instant, the algorithm needs to store all previous data samples.
In the context of kernel methods, this set of stored data is usually called ``\emph{dictionary}''.
By contrast, to distinguish it from the ``\emph{representation dictionary}'' used in the context of sparse representation,
we use the term ``\emph{feature space profile}'', or ``\emph{profile}'' to refer to
these data samples along with all corresponding matrixes and vectors that collectively represent the representation dictionary.
In practice, redundancy among the input data makes it possible to reduce the size of this profile,
at the cost of a negligible effect on the performance of the model.
In the context of kernel methods, this procedure is termed ``\emph{sparsification}'',
but we call this procedure ``\emph{profile abstraction}'', or ``\emph{abstraction}''.
Abstraction is essentially an active data selection problem in which the aim is to
select the most informative or the most salient data points.
This can be accomplished based on two different approaches, namely ``\emph{growing}'' and ``\emph{pruning}''.

\subsubsection{Profile Growing}\label{SubSubGrow}

In the growing procedure, the algorithm starts with an empty, or an initially filled, profile and
gradually adds important data samples into the profile, as iterations go on and new learning vectors arrive.
This is generally carried out greedily by selecting, based on a measure, the most informative data
and discarding the others.
Note that, when any sample data is included in the profile,
the definition of most of the previously defined variables should be revised accordingly.
The update equations for this step were derived in subsection \ref{SubSecAdd}.
Furthermore, the set $\Omega_i$ used in defining the $\xi$ value in \Eq{eqXi_i}
is the set of forgetting factors that correspond to all input data samples that have been selected in this step.

\emph{Coherency} and \emph{representation error} are two criteria that have been widely used,
in the context of kernel learning, for profile abstraction \cite{BibLiu10}.
We use the coherency criterion based on which the new data sample $\mathbf{x}$ is added to the profile,
if the normalized coherency of $\phivec$ with the most coherent data sample within the profile is small enough.
More specifically, the new data is accepted to be informative if:
\begin{equation}\label{eqCoh}
    max_j |\frac{k_j}{\sqrt{\sigma_x^2 K_{jj}}}| < \delta,
\end{equation}
where $\delta$ is a pre-set threshold.
A closely related criterion is the \emph{projection} measure \cite{BibBrien16},
based on which the new data sample is added to the profile if the projection of $\phivec$ into the space spanned by the columns of $\Phi_i$ is small enough.

\subsubsection{Profile Pruning}\label{SubSubPrun}

The pruning procedure is based on excluding less important elements from the profile,
to restrict the size of the profile and hence harness the complexity of the algorithm.
This can be done by removing a properly-selected mini-batch of $M$ data samples,
when the size of the profile exceeds the pre-set threshold $L_{max}$ so as the size of profile reduces to $L_{max} - M$.
Again, the indices of the data sample that should be removed are stored in vector $\mathbf{m}$.
In our problem, removing the stored input vectors necessitates updating some matrixes,
as described in subsection \ref{SubSecRmv}. However, pruning does not affect the set $\Omega_i$;
as this set corresponds to the input samples that have been included in the profile.
This special definition is necessary for deriving \Eq{eqC_h}.

One idea for selecting the least important input sample is to discard data samples in the order of their \emph{entrance}.
Specifically, the effect of the first $M$ data samples, that are the oldest input vectors in $X_{i}$, can be removed.
A more reasonable idea is to discard data samples with the least \emph{contribution} in representation.
In the proposed approach, this contribution should be calculated in the feature space.
At each time instant, the input data samples mapped into the feature space, i.e. $\Phi_i$,
is approximated based on the profile dictionary $\Dict_i$ using the coefficient matrix $W_i$ as:
\begin{eqnarray}\label{eqAPhi1}
    \hat{\Phi}_i = D_i W_i = (\Phi_i U_i^T) W_i.
\end{eqnarray}
Defining $B_i = U_i^T W_i \in \mathbb{R}^{L \times L}$, this gives:
\begin{eqnarray}\label{eqAPhi2}
    \hat{\Phi}_i = D_i W_i = \Phi_i B_i.
\end{eqnarray}
Column $j$ of $B_i$ thus gives the coefficients for the linear combination of the mapped data samples within the profile,
i.e. the columns in $\Phi_i$, as used in the approximation of the column $j$ in $\Phi_i$.
In other words, each row of $B_i$ represents the contribution weights of the corresponding data sample in representation of all retained data samples.
Based on this criterion, denoting the $j$th row of the matrix $B_i$ by $\breve{b}_j$, the indexes of the candidate data samples is found as those with the minimum value of $|| \breve{b}_j^T||_2$.
But, to consider also the order of the entrance, we decided to select the $M$ data samples based on the proposed contribution criterion,
only from the first half of the data samples within the profile.

However, when a data sample is discarded from the profile, zeroing out the corresponding coefficient vector of $W_{i+1}$
may render $\big(\lamvec_m^{-1} - \mathbf{w}_m^T \mathbf{u}_m \big)$ singular.
Note that, unlike the $\alpha$ parameter defined in \eqref{eqalpha},
the denominator of $\alpha_m$ in \eqref{eqalpham} may be singular.
In this situation, the matrix $W_{i+1} \hat{\Lambda} W_{i+1}^T$ is singular and hence it does not have an inverse.
Moreover, as the coefficient vectors are generally sparse,
removing $\mathbf{w}$ may cause some rows of the matrix $W_{i+1}$ to be all-zero
and hence render the matrix $W_{i+1} \hat{\Lambda} W_{i+1}^T$ non-invertible.
Therefore, to make certain that removing the coefficients vectors
does not cause the $W_{i+1}$ matrix to become singular, two conditions must be checked, before removing:
$\big(\lamvec_m^{-1} - \mathbf{w}_m^T \mathbf{u}_m \big)$ is non-singular
and all rows of $W_{i+1}$ at least have a non-zero element.
In other words,  we decide to select the first $M$ coefficient vectors, based on the proposed contribution criterion,
for which these two conditions are satisfied.

\subsection{Dictionary Normalization} \label{SubSecNorm}
In some situations, it is preferable to normalize the dictionary so that all atoms have unit $\ell_2-$norm.
This restrains the approximation from trivial solutions.
However, the profile update steps do not preserve the norm of the dictionary columns
and hence it may be necessary to re-normalize the dictionary, even if periodically.
The virtual dictionary $D$ can be simply normalized by rescaling the Gram matrix $\Psi$.
Defining a diagonal matrix $S$ whose $j$th diagonal element is the square root of the $j$th diagonal element of the matrix $\Psi$,
this rescaling can be formulated as:
\begin{equation}\label{eqPhin}
  \Psi_n = S^{-1}\Psi S^{-1},
\end{equation}
where $S^{-1}$ is the inverse of $S$.
This normalization necessitates rescaling some other matrixes, in particular $W$, $C$, and $U$ matrixes must be rescaled as:
\begin{equation}\label{eqScale}
  W_n = S W,
  \quad C_n = S^{-1} C S^{-1}
  \quad and \quad U_n = S^{-1} U.
\end{equation}

\subsection{Final Algorithm}\label{SubSecFnl}
In the proposed algorithm, the dictionary is represented by the profile that is the collection of
$X_i$, $K_i$, $W_i$, $C_i$, $U_i$, and $\Psi_i$  matrixes and $\lamvec_i$ vector.
The algorithm starts with initializing the profile.
Assuming that at start $L=Q$, the columns of $X_i$ can be filled randomly or, if available, with $Q$ data samples
and then $K_i$ is calculated using the kernel function.
By initializing $X_i$ in this way, $W_i$, $C_i$, and $U_i$ are initialized with $Q \times Q$ identity matrixes
and $\lamvec_i$ is formed with the $Q$ first values of the $\lambda$ parameter, that we assume to be all $1$.
After initializing, the profile is sequentially updated at each iteration,
according to the remaining steps summarized in Algorithm \ref{Alg}.

\begin{algorithm}[H]\label{Alg}
    \caption{KRLS DL Algorithm}
    \SetAlgoLined
    \setcounter{AlgoLine}{-1}
    \KwIn{Training data samples $\mathbf{x} \in \mathbb{R}^N$, that are sequentially fed into the algorithm, in mini-batches of $M$ vectors}
    \KwOut{The updated profile, represented by $X_i$, $K_i$, $W_i$, $C_i$, $U_i$, $\Psi_i$, and $\lamvec_i$}
    \textbf{Initialization:} Initialize the profile, as described in subsection \ref{SubSecFnl}. While new data is available, iterate over the next steps.\\
    \textbf{New Data:} Get the new $M$ data samples $\mathbf{x}$ and check for the informativeness of each data sample independently,
        e.g. using \eqref{eqCoh}.
        If the informativeness of at least one data sample is satisfied, discard the non-informative vectors and go to the next step;
        otherwise, go to Step 1 on the next iteration.\\
    \textbf{SA:}       Calculate $\mathbf{w}$ by SA, e.g. using the KORMP algorithm.\\
    \textbf{Profile Growing:} Grow the profile, using equations
        \eqref{eqlam_i1}, \eqref{eqXW_i1}, \eqref{eqC_i1}, \eqref{eqK_i1}, \eqref{eqUi1}, and \eqref{eqPsi2}.\\
    \textbf{Profile Pruning:} If the profile size is greater than $L_{max}$, prune the profile, using equations
        \eqref{eqChat}, \eqref{eqUh}, and \eqref{eqPhi_h}, and then discard the columns $\mathbf{m}$ of $X_i$, $W_i$, $K_i$, and
        $\hat{U}$, the rows $\mathbf{m}$ of $K_i$ , as well as the elements $\mathbf{m}$ of $\lamvec_i$.\\
    \textbf{Normalization:} Normalize the dictionary based on the procedure described in subsection \ref{SubSecNorm}, if necessary, but normally not.
\end{algorithm}

It is worth noting that the proposed algorithm still involves matrix inversion in equations \eqref{eqalpha} and \eqref{eqalpham}. But now, the matrixes that must be inverted are of size $M \times M$, which can be selected very small. More importantly, for the case $M=1$, which is quite a usual case in practice, matrix inversion is replaced by division.
With regard to the profile abstraction, our tests showed that
by just checking the informativeness of the new data vectors, based on the growing procedure,
the size of the profile may grow well beyond a reasonable bound.
Therefore, it may be necessary to implement this algorithm in a fixed-budget fashion,
i.e. restricting the size of the profile to a predefined limit, say $L_{max}$, based on the pruning procedure.
When a new batch of input vectors are included in the profile or old vectors are excluded from the profile,
corresponding matrixes are updated based on the formulas derived in
subsections \ref{SubSecAdd} and \ref{SubSecRmv}, respectively.
Moreover, since in practice, the dictionary changes slowly over successive iterations,
and as an effect of the inclusion of the regularization term in the problem \eqref{eqLS1},
one can skip the last step, i.e., normalization, for most iterations, and employ it periodically.
In the simulation tests reported in the next section, we decided to check for the necessity of this step every time the profile is pruned.

\section{Simulation Tests}\label{SecTest}
To evaluate the performance of the proposed online dictionary learning algorithm,
some numerical tests were conducted for the KSR-based classification task
in four different applications.
This evaluation was carried out in comparison to existing online KDL algorithms.
To the best of our knowledge, hitherto three online KDL algorithms have been proposed for general kernels
\cite{BibBrien16}, \cite{BibLiu16} and \cite{BibLee16}, whose algorithms are denoted as sKDL, OKDLRS and OKDLFB, respectively.
Furthermore, the batch KMOD algorithm proposed in \cite{BibNguyen13} is considered as the benchmark.
The main results of these tests are reported in this section.
It is noted that batch DL algorithms are not alternatives to online DL algorithms.
As explained in section \ref{SecInt}, the computational cost and storage cost of batch algorithms increase prohibitively, as size of the training set increases.
This can also be seen from the results reported later in Table \ref{TabTime} for moderately sized datasets used in the experiments.

\subsection{Datasets}\label{SubSecData}
The simulation tests were conducted on the following datasets:

\textbf{USPS:} United States Postal Service (USPS) digit recognition dataset contains $16 \times 16-$pixel images of
handwritten digits from $0$ to $9$ in $8-$bit grayscale, created by scanning digits on the envelopes.
Therefore, a ten-class classification task is defined for digit recognition.
In the version used in the current study \footnote{\url{https://cs.nyu.edu/~roweis/data/usps_all.mat}},
each class contains $1100$ images and all images are first vectorized to $256-$dimensional data samples, i.e. $N=256$.

\textbf{ISOLET:} ISOlated LETter (ISOLET) recognition dataset contains speech-specific features extracted from spoken letters.
To generate this dataset, $150$ subjects spoke the name of each letter of the English alphabet twice;
that is there are $300$ samples for each letter.
$617$ features, including spectral, contour, sonorant, pre-sonorant, and post-sonorant features, were extracted from each utterance;
therefore in this case $N=617$.
All feature values are continuous, real-valued attributes scaled into the range $-1.0$ to $1.0$.
This dataset \footnote{\url{https://datahub.io/machine-learning/isolet/r/isolet.csv}}
is used for spoken letter recognition which is a $26-$class classification task.

\textbf{EEG:} EEG in schizophrenia dataset \cite{BibRepod17} comprised $14$ patients with paranoid schizophrenia and $14$ healthy controls.
Therefore,  the aimed problem is a $2-$class classification task.
Data were acquired with the sampling frequency of $250 Hz$ using the standard $10-20$ EEG montage with $19$ EEG channels.
The signals of healthy and patient cases are of length $925$ and $845$ seconds, respectively, but only the first $169$ seconds of signals, in both classes, were used in the current study.
Each $19-$channel signal is divided into non-overlapping segments of $1-$second length, and each segment is regarded as an independent data sample.
Hence, each class contain $2366$ data samples.
Four frequency features, known as delta ($0-4 Hz$), theta ($4-8 Hz$), alpha ($8-13 Hz$), and beta ($13-31 Hz$), were extracted from each channel.
This procedure results in $4$ features over each channel, i.e., totally $N=76$ features are extracted from each data sample.

\textbf{DistNet} One feeder named “Persian Gulf” from the “Malayer $1$” distribution substation of the “Hamedan power distribution system” in Iran, which is a critical feeder, was simulated to generate simulated data for fault detection. The simulation was done in Electromagnetic Transient Program (EMTP) software under various fault conditions and factors. In addition to the no-fault condition, different short-circuit faults, including Single-Line-to-Ground (SLG), Line-to-Line (LL), Double-Line-to-Ground (LLG), and Triple-Line (LLL), were considered. $400$ data samples were generated for each condition, where for each case, the main factors of the grid and fault (including the fault location, the load value of each consumer, the fluctuation of the voltage of the main source, the inception angle, and the impedance of the fault) were assigned randomly, within reasonable ranges. One cycle ($201$ samples) of the three-phase voltage and current signals at the beginning of the feeder was recorded with the sampling frequency of $10 KHz$. These signals were concatenated and are used for this study. In summary, generated data samples are of $N=1206$ dimensions and are employed for a $5-$class classification task.

\subsection{Test Scenario}\label{SubSecScenario}
All evaluations are performed using $5-$fold cross-validation.
One dictionary is trained for each class and each test sample is assigned to the class
whose dictionary results in the minimum representation error.
For online DL learning algorithms, training samples, i.e. all samples within other $4$ groups over each fold,
are exposed to the algorithm in mini-batches of $M$ samples.
By receiving a mini-batch of data from each class, the corresponding profile is updated.
The learning procedure is proceeding for a predefined number of batches.
If the total number of available batches of training samples is less than that number,
the learning procedure sequentially iterates over the previously seen data vectors, for some epochs.
As the profiles are sequentially trained based on the training data samples,
the updated dictionaries are intermittently used to classify all test samples, i.e. the samples in the hold-out group.
The total number of tests is set to $20$.
Ideally, by increasing the number of training batches and exposing more data samples,
the classification accuracy should increase and reach the accuracy obtained by the batch KMOD DL algorithm.

For all tests reported herein, a quadratic polynomial kernel is used.
The forgetting factor $\lambda$ linearly increases from $0.98$ to $1$ for the first $80\%$ of batches
and is then kept constant for the remaining batches.
Moreover, the number of dictionary atoms and the maximum size of the profile
are set to $Q=30$ and $L_{max}=200$, respectively.
The batch size for both growing and pruning steps were set to $10$,
and $\gamma$ regularization parameter is set to $0.1$.
To provide a fair comparison among different DL algorithms,
the test samples were always approximated using the KORMP algorithm \cite{BibSkretting08}
with sparsity level $s=5$, regardless of the approximation algorithm proposed in the DL algorithm.
 Each algorithm is implemented with its own parameter values that result in higher accuracy.
As an indication of the computational complexity of the algorithms, the average time necessary to train one dictionary over each mini-batch is also reported.
It should be noted that for the online DL algorithms the training time depends on the profile size
that gradually increases to $L_{max}$, as mini-batches arrive.
All simulation tests were conducted using {\matlab}  $2019b$ in $64$bit Windows operating system
on a $3.2 GHz$ Intel CPU and $16 GB$ memory while only one core is enabled
\footnote{Simulation codes are available at \url{https://github.com/G-Alipoor/KRLS_DLA}.}.

\subsection{Results}\label{SubSecRes}
\begin{figure}[h]\centering
    \begin{subfigure}[!ht]{0.49\textwidth}\centering
      \centering
      \includegraphics[width=1\textwidth]{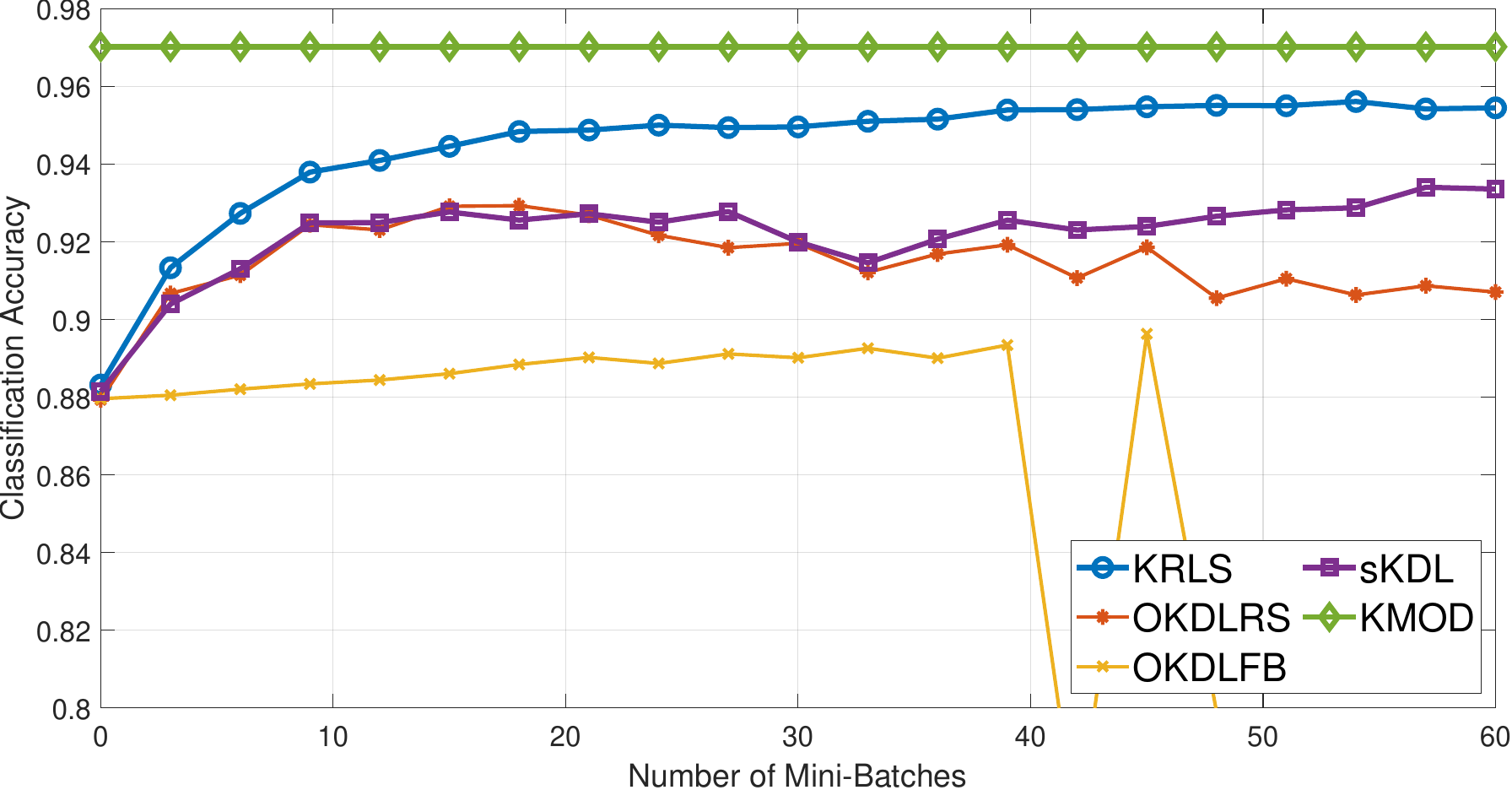}
      \caption{USPS Dataset}
    \end{subfigure}
    \begin{subfigure}[!ht]{0.49\textwidth}\centering
      \centering
      \includegraphics[width=1\textwidth]{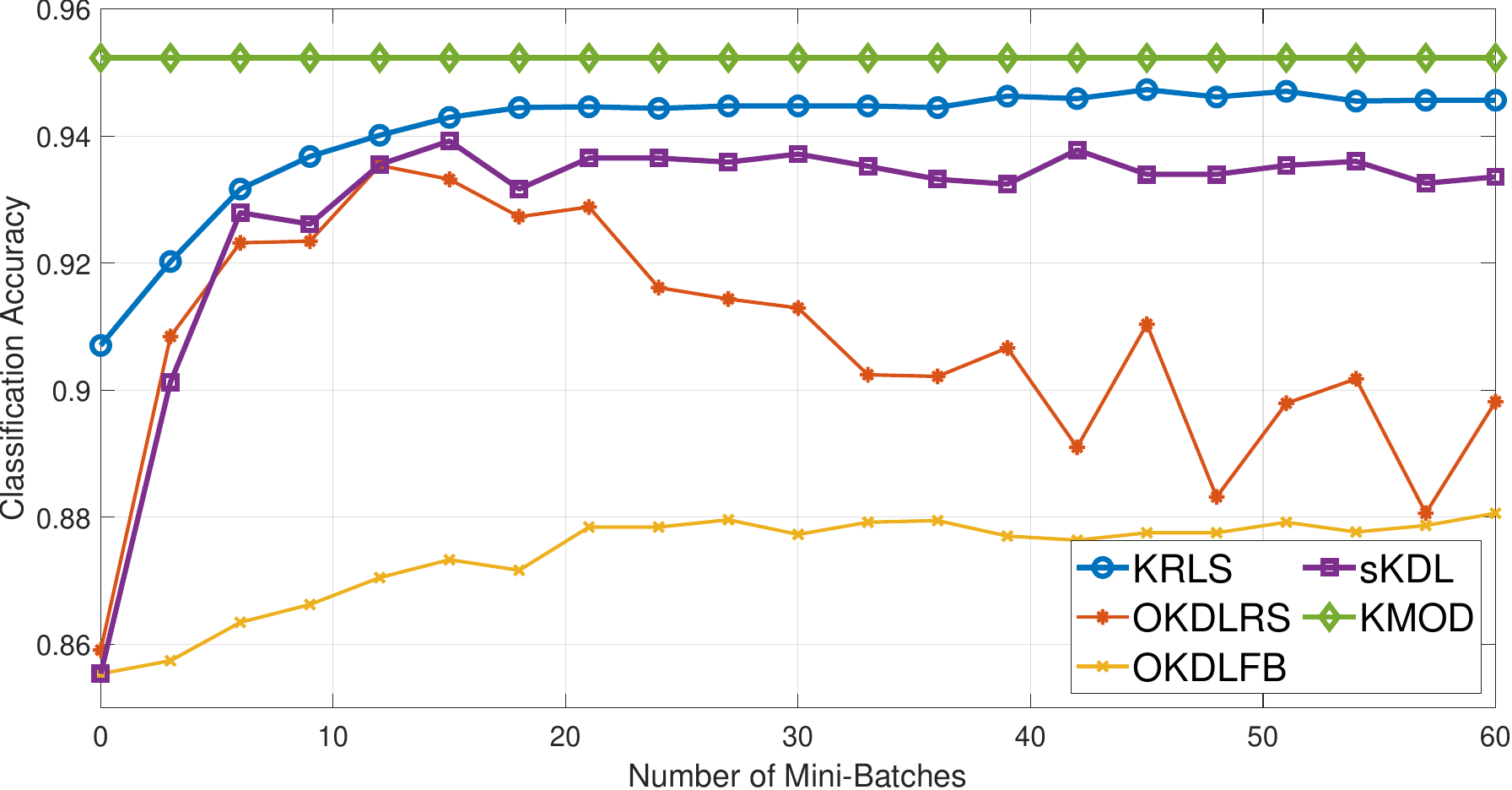}
      \caption{ISOLET Dataset}
    \end{subfigure}
    \begin{subfigure}[!ht]{0.49\textwidth}\centering
      \centering
      \includegraphics[width=1\textwidth]{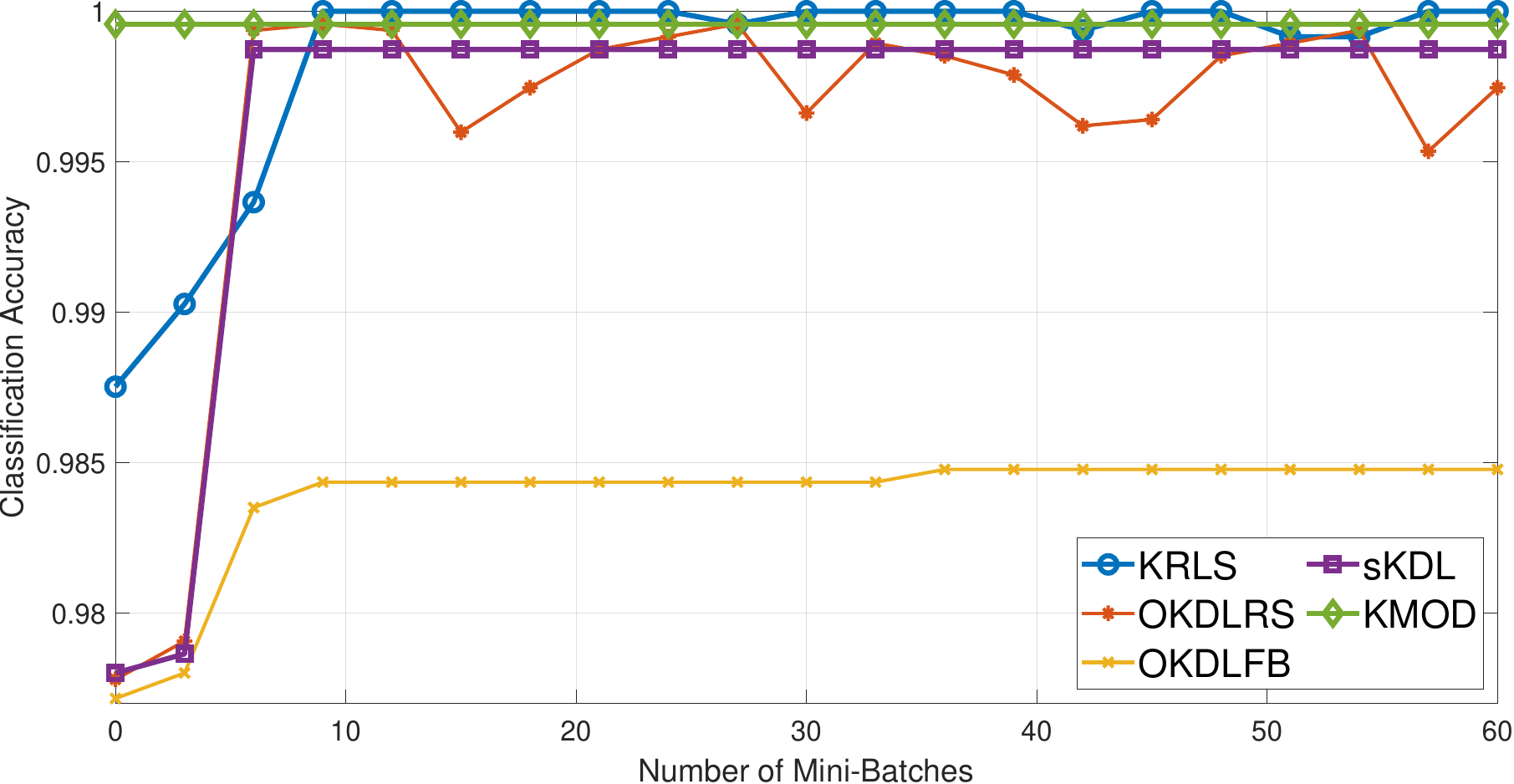}
      \caption{EEG Dataset}
    \end{subfigure}
    \begin{subfigure}[!ht]{0.49\textwidth}\centering
      \centering
      \includegraphics[width=1\textwidth]{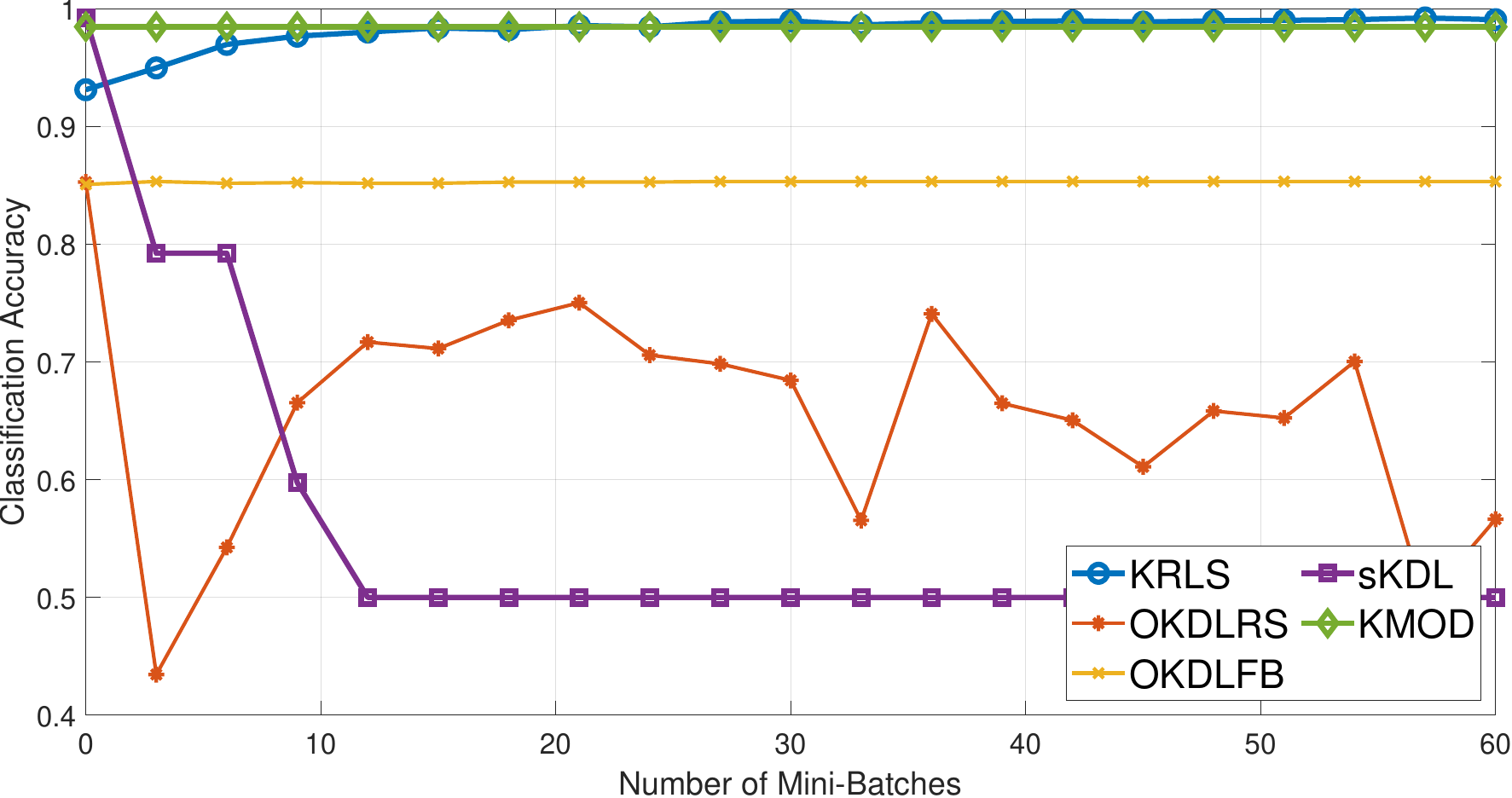}
      \caption{DistNet Dataset}
    \end{subfigure}
    \caption{The averaged classification accuracy of DL algorithms}
    \label{FigComp}
\end{figure}

Classification accuracies averaged over $k=5$ folds as a function of the number of mini-batches for four datasets are shown in Figure \ref{FigComp},
where the performance of the proposed algorithm is compared with that of all other existing KDL algorithms.
The averaged accuracy achieved by the batch KMOD algorithm is also reported; this can be considered as a benchmark.
Furthermore, the average training time of one dictionary over each mini-batch for the online kernel algorithms,
along with the average execution time of each iteration of the batch algorithm for one dictionary, are tabulated in Table  \ref{TabTime}.
The average time of the two main tasks of the training procedure of the proposed algorithm is reported in this table.
The growing task represents steps $2$ and $3$ of the algorithm, i.e., SA and profile growing,
while the pruning task accounts for steps $4$ and $5$, i.e., profile pruning and normalization.
The total average training time of the KRLS algorithm hence equals the sum of these two values.
The sKDL algorithm proposed in \cite{BibBrien16} uses the projection criterion to select the most informative incoming data samples for inclusion in the dictionary,
similar to the growing step of the KRLS algorithm.
This criterion is not satisfied for almost all data samples of the DistNet dataset; hence this algorithm ignores these samples.
Consequently, the accuracy curve does not change for most mini-batches (since the dictionary is not updated over these batches),
and the training procedure is carried out in a relatively short time.
Our examination showed that using other growing criteria, apart from what was proposed in the original algorithm, does not improve the result over this dataset.

These results clearly show the outperformance of the proposed algorithm.
In particular, during the training progress, the proposed algorithm is always of higher classification accuracy.
Moreover, the accuracy achieved by the proposed algorithm almost monotonically increases toward that of the batch algorithm;
this observation shows the high scalability of the KRLS DL algorithm.
This improvement is achieved despite a smaller execution time.
The decrease in the training time is an indication of the smaller computational complexity of the KRLS DL algorithm
that is in turn a direct effect of the efficient derivation of the algorithm.

\begin{table} [h] \centering
    \caption{Average training time (in milliseconds) necessary to train one dictionary over each mini-batch}
    \resizebox{\textwidth}{!}{
    \begin{tabular}{l l c c c c c c}
        \hline
        \multicolumn{2}{l}{\multirow{2}{*}{Algorithm}} & \multirow{2}{*}{OKDLRS} & \multirow{2}{*}{OKDLFB} & \multirow{2}{*}{sKDL} & \multirow{2}{*}{KMOD} &
        \multicolumn{2}{c}{KRLS}  \\
                                    &                   &                         &                         &                       &                       &
         Growing & Pruning \\
        \hline
        \multirow{4}{*}{\begin{turn}{90} Dataset \end{turn}} &  USPS   &  2856.6 &  141.5  &  87.2  &  4651.3 &  2.0  & 2.3 \\
                                 & ISOLET  & 2868.3  &  117.5  &  37.3  &  114.1  &  2.3  & 2.5 \\
                                 &   EEG   &  2910.5 &  5.1    &  0.7   & 69860   &  1.8  & 2.4 \\
                                 & DistNet & 354.3   &  55.2   &   .5   &  174.9  & 5.4   & 5.1  \\
        \hline
    \end{tabular}}
    \label{TabTime}
\end{table}

\begin{figure}[h]\centering
    \begin{subfigure}[!ht]{0.49\textwidth}\centering
      \centering
      \includegraphics[width=1\textwidth]{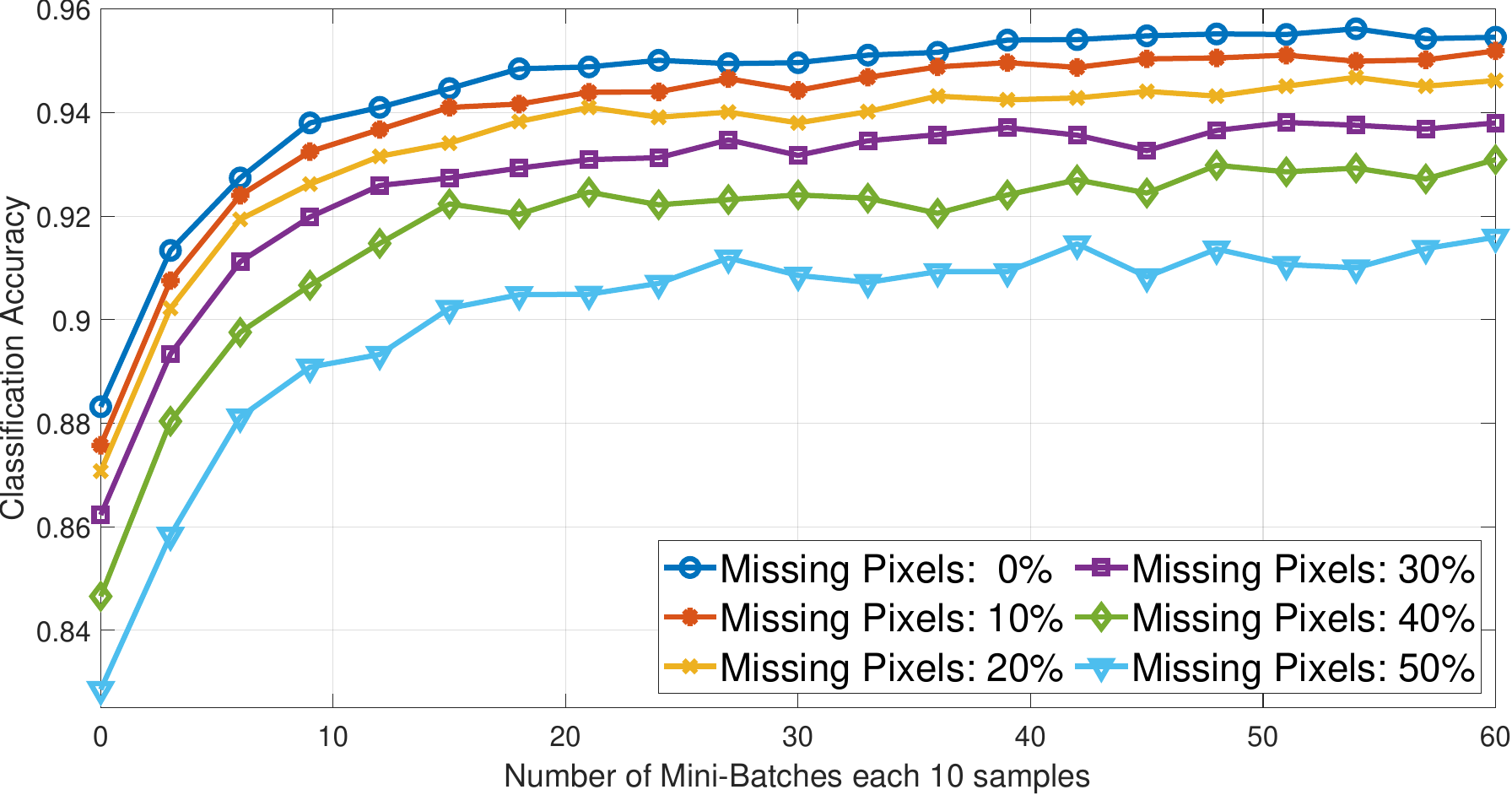}
      \caption{USPS Dataset}
    \end{subfigure}
    \begin{subfigure}[!ht]{0.49\textwidth}\centering
      \centering
      \includegraphics[width=1\textwidth]{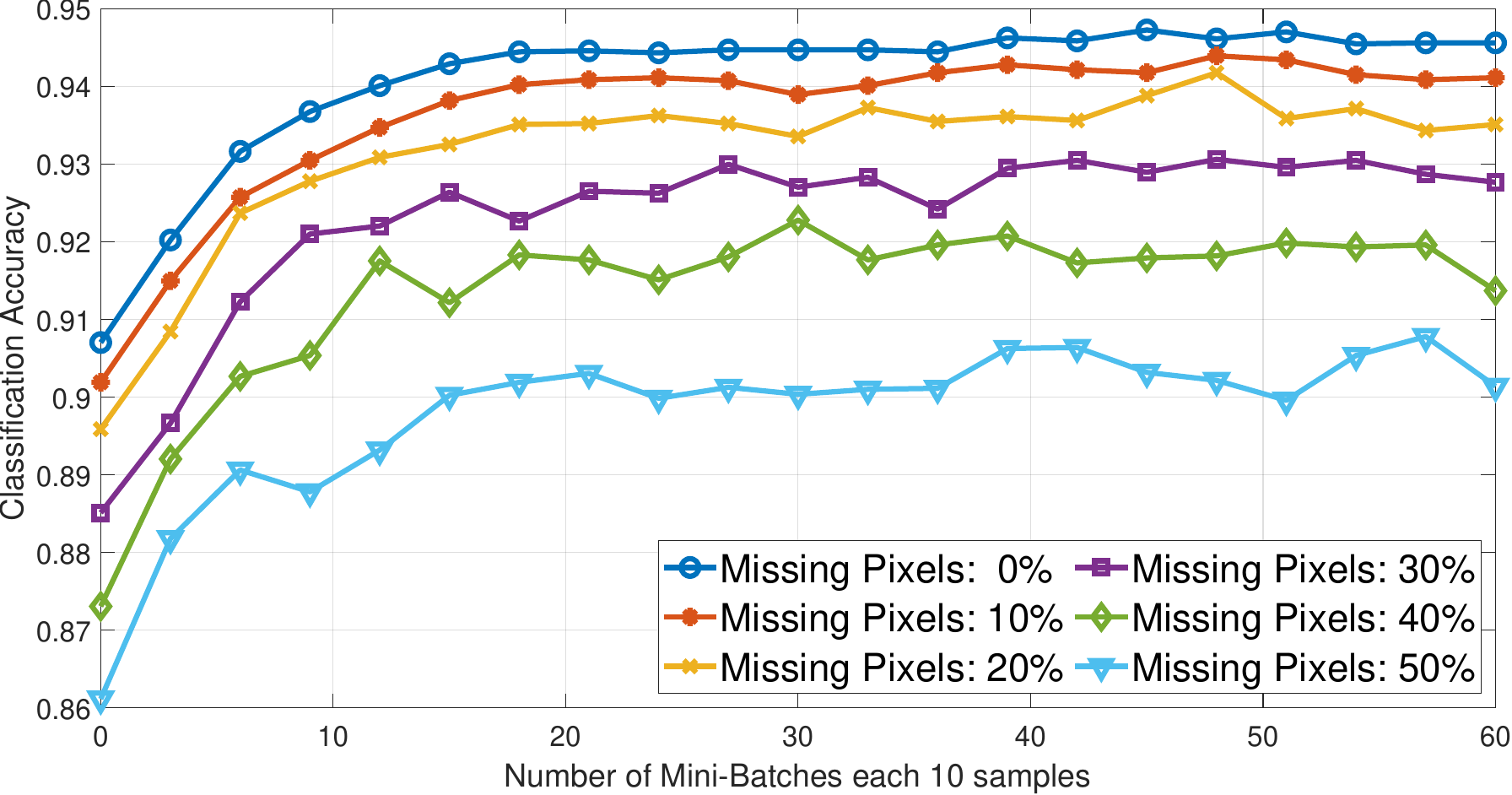}
      \caption{ISOLET Dataset}
    \end{subfigure}
    \begin{subfigure}[!ht]{0.49\textwidth}\centering
      \centering
      \includegraphics[width=1\textwidth]{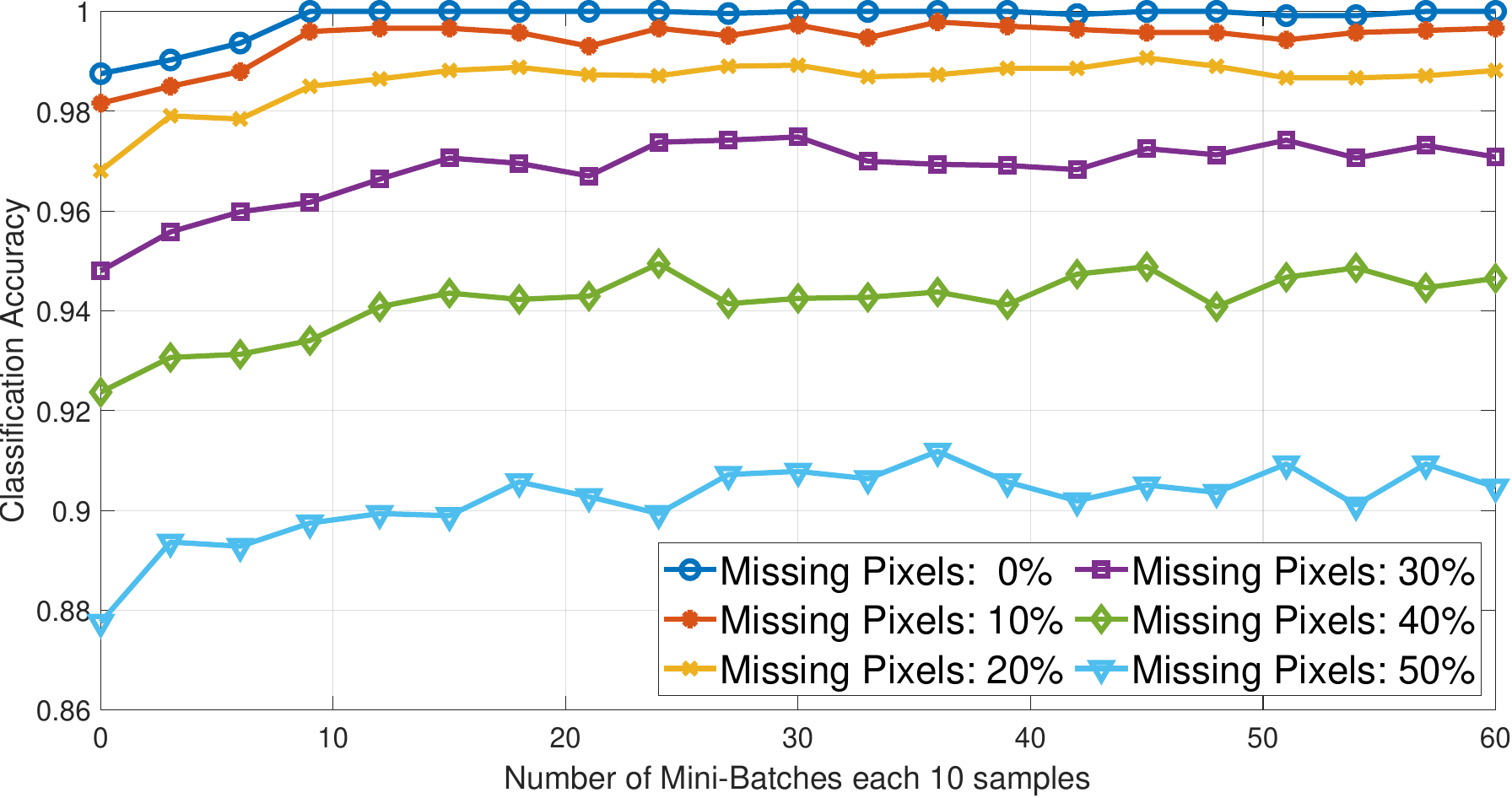}
      \caption{EEG Dataset}
    \end{subfigure}
    \begin{subfigure}[!ht]{0.49\textwidth}\centering
      \centering
      \includegraphics[width=1\textwidth]{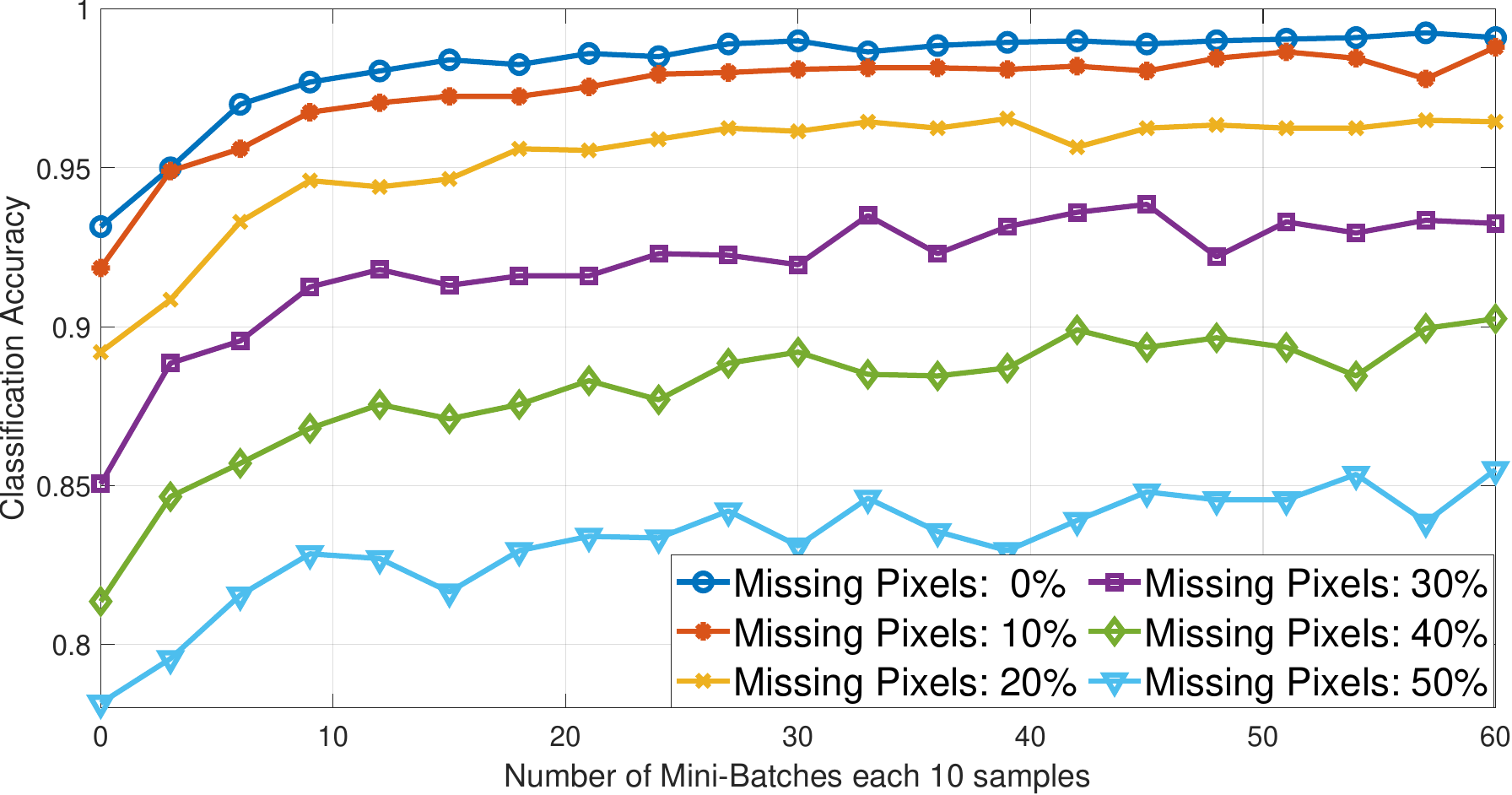}
      \caption{DistNet Dataset}
    \end{subfigure}
    \caption{Effect of missing pixels on the classification accuracy of the KRLS DL algorithm}
    \label{FigNoise}
\end{figure}

In another test, the robustness of the dictionaries trained with the proposed algorithm is evaluated.
For this aim, a randomly selected portion of the data points of all test samples are removed by replacing their values with $0$.
It should be noted that this destruction is only applied to the test data samples,
that is dictionaries are trained with the un-changed data samples.
The averaged classification accuracies for different percentages of missing data points are summarized in figure \ref{FigNoise}.
These results show the robustness of the dictionaries trained with the KRLS DL algorithm against noise,
even in severe cases when most data points are removed.

\section{Conclusion} \label{SecCon}
Online dictionary learning algorithms have appealing properties in many applications where it is not practically possible to utilize batch algorithms. An online kernel dictionary learning algorithm was developed in this work whose objective is to sequentially optimize a dictionary in a feature space induced by a Mercer kernel that minimizes a weighted recursive least squares function of the sparse representation error in the same feature space. This recursive solution was derived using the matrix inversion lemma. The proposed algorithm is an online version of the KMOD batch dictionary learning algorithm and at the same time can be viewed as an extension of the RLS dictionary learning algorithm to kernel space. In spite of the possibly apparent similarity, this work addressed an absolutely different problem with its own concepts and challenges. Moreover, the derived algorithm is the first LS-based kernel dictionary learning algorithm that inherits the good property of a higher convergence rate from the LS approach.

An efficient implementation of the solution is also derived which makes the implementation only-matrix-by-vector for batch size of $1$ sample and thus the final algorithm is of considerably lower computational complexity. Experimental tests conducted on four practical datasets in quite different applications, i.e., image recognition, sound classification, disorder diagnosis in medical science, and fault detection in power distribution networks, showed the outperformance of the proposed algorithm, in terms of both classification accuracy and computational burden. Overall, we think that the proposed algorithm is a considerable progress in the field of online kernel dictionary learning algorithms.

\appendix
\section{Appendix: Some Mathematical Derivations}
Detailed derivations for equations \eqref{eqUi1}, \eqref{eqDi1}, \eqref{eqPsi2}, \eqref{eqChat}, and \eqref{eqDh1}-\eqref{eqPhi_h} are provided in this appendix.

\subsection{Equation \eqref{eqUi1}} \label{appendix:gU}
\begin{eqnarray}
  U_{i+1} & = & C_{i+1} W_{i+1} \Lambda_{i+1} \nonumber \\
          & = & \lambda^{-1} \, \big( C_i \, - \, \mathbf{u} \alpha \mathbf{u}^T \big)
                  [W_i \, \lambda \Lambda_i, \, \mathbf{w}] \nonumber \\
          & = & \lambda^{-1} \, \big[ C_i W_i \lambda \Lambda_i -
                  \mathbf{u} \alpha \mathbf{u}^T W_i \lambda \Lambda_i,
                  \  C_i \mathbf{w} - \mathbf{u} \alpha \mathbf{u}^T \mathbf{w} \big]
                  \nonumber \\
          & = & \lambda^{-1} \big[ \lambda U_i -
                     \mathbf{u} \alpha \, \mathbf{u}^T W_i \lambda \Lambda_i, \
                     \mathbf{u} - \mathbf{u} (I_M - \lambda \alpha) \, \big] \nonumber \\
  U_{i+1} & = & \Big[ \, U_i - \mathbf{u} \alpha \, \mathbf{v}^T,
                           \, \mathbf{u} \alpha \, \Big],  \nonumber
\end{eqnarray}

\subsection{Equation \eqref{eqDi1}} \label{appendix:gD}
\begin{eqnarray}
  \Dict_{i+1} & = & \Phi_{i+1} U_{i+1}^T
           =  [ \Phi_i, \phivec ]
           \left[ \begin{array}{c}   U_i^T - \mathbf{v} \alpha \mathbf{u}^T \\
                  \alpha \mathbf{u}^T   \end{array} \right]  \nonumber \\
          & = & \Phi_i U_i^T - \Phi_i \mathbf{v} \alpha \mathbf{u}^T
                     + \phivec \alpha \mathbf{u}^T
           =  \Dict_i + (\phivec - \Phi_i \mathbf{v}) (\alpha \mathbf{u}^T) \nonumber \\
          & = & \Dict_i + \mathbf{r} \alpha \mathbf{u}^T, \nonumber
\end{eqnarray}

\subsection{Equation \eqref{eqPsi2}} \label{appendix:gPsi}
\begin{eqnarray}
  \Psi_{i+1} & = & D_{i+1}^T D_{i+1} = U_{i+1} \, K_{i+1} \, U_{i+1}^T \nonumber \\
       & = & [ \UL, \mathbf{u} \alpha ]
       \left[ \begin{array}{cc}   K_i & \mathbf{k} \\
                  \mathbf{k}^T & \sigma_{\mathbf{x}}^2   \end{array} \right]
       \left[ \begin{array}{c}  \ULT \\ \alpha \mathbf{u}^T  \end{array} \right] \nonumber \\
       & = & (\UL) K_i (\ULT) \nonumber \\
       &   & + (\UL) \mathbf{k} \alpha \mathbf{u}^T + \mathbf{u} \alpha \mathbf{k}^T (\ULT)
             + \mathbf{u} \alpha \sigma_{\mathbf{x}}^2 \alpha \mathbf{u}^T \nonumber \\
       & = & U_i K_i U_i^T
             - \mathbf{u} \alpha \mathbf{v}^T K_i U_i^T
             - U_i K_i \mathbf{v} \alpha \mathbf{u}^T
             + \mathbf{u} \alpha \mathbf{v}^T K_i \mathbf{v} \alpha \mathbf{u}^T \nonumber \\
       &   & + \, U_i \mathbf{k} \alpha \mathbf{u}^T
             - \mathbf{u} \alpha \mathbf{v}^T \mathbf{k} \alpha \mathbf{u}^T
             + \mathbf{u} \alpha \mathbf{k}^T U_i^T
             - \mathbf{u} \alpha \mathbf{k}^T \mathbf{v} \alpha \mathbf{u}^T
             + \mathbf{u} \alpha \sigma_{\mathbf{x}}^2 \alpha \mathbf{u}^T \nonumber \\
       & = & \Psi_i + \mathbf{u} \alpha (\mathbf{k}^T - \mathbf{v}^T K_i) U_i^T
             + U_i (\mathbf{k} - K_i \mathbf{v}) \alpha \mathbf{u}^T \nonumber \\
       &   & + \, \mathbf{u} \alpha (\mathbf{v}^T K_i \mathbf{v}
               - \mathbf{v}^T \mathbf{k} - \mathbf{k}^T \mathbf{v}
               + \sigma_{\mathbf{x}}^2 ) \alpha \mathbf{u}^T \nonumber \\
  \Psi_{i+1} & = & \Psi_i + \mathbf{u} \alpha \mathbf{\tilde{u}}^T
             + \mathbf{\tilde{u}} \alpha \mathbf{u}^T
             + \mathbf{u} \alpha (\mathbf{v}^T K_i \mathbf{v}
               - \mathbf{v}^T \mathbf{k} - \mathbf{k}^T \mathbf{v}
               + \sigma_{\mathbf{x}}^2 ) \alpha \mathbf{u}^T, \nonumber
\end{eqnarray}

\subsection{Equation \eqref{eqChat}} \label{appendix:pC}
\begin{eqnarray}
  \hat{C} & = & C_i - C_i  \mathbf{w}_m \lamvec_m^{0.5}
             \Big(-I_M +  \lamvec_m^{0.5} \mathbf{w}_m^T C_i \mathbf{w}_m \lamvec_m^{0.5} \Big)^{-1}
             \lamvec_m^{0.5} \mathbf{w}_m^T C_i \nonumber \\
          & = & C_i + C_i  \mathbf{w}_m \Big(\lamvec_m^{-1} - \mathbf{w}_m^T C_i \mathbf{w}_m \Big)^{-1} \mathbf{w}_m^T C_i \nonumber \\
  \hat{C} & = & C_i + \mathbf{u}_m \alpha_m \mathbf{u}_m^T, \nonumber
\end{eqnarray}

\subsection{Equation \eqref{eqDh1}} \label{appendix:pD1}
\begin{eqnarray}
  \hat{D} = \Phi_i \hat{U}^T & = & \Phi_i[U_i^T - U_{\downarrow m}^T + \mathbf{v}_m \alpha_m \mathbf{u}_m^T] \nonumber \\
                             & = & \Phi_i U_i^T - \Phi_i U_{\downarrow m}^T + \Phi_i \mathbf{v}_m \alpha_m \mathbf{u}_m^T \nonumber \\
                             & = & D_i - [\mathbf{\phi}_m \lamvec_m  - \Phi_i \mathbf{v}_m \alpha_m ]\mathbf{u}_m^T. \nonumber
\end{eqnarray}

\subsection{Equation \eqref{Pshi_v}} \label{appendix:phiv}
\begin{eqnarray}
  \Phi_i \mathbf{v}_m = \Phi_i \hat{\Lambda} W_i^T \mathbf{u}_m
        & = & \Phi_i \hat{\Lambda} W_i^T C_i \mathbf{w}_m \nonumber \\
        & = & \Phi_i \Lambda_i W_i^T C_i \mathbf{w}_m - \Phi_i \Lambda_{\downarrow m} W_i^T C_i \mathbf{w}_m \nonumber \\
        & = & \hat{\phivec}_m - \phivec_m (\lamvec_m \mathbf{u}_m^T \mathbf{w}_m) \nonumber \\
        & = & \hat{\phivec}_m - \phivec_m + \phivec_m \lamvec_m \alpha_m^{-1}. \nonumber
\end{eqnarray}

\subsection{Equation \eqref{eqDh2}} \label{appendix:pD2}
\begin{eqnarray}
  \hat{D} & = & D_i - [\phivec_m \lamvec_m - \hat{\phivec}_m \alpha_m
                 + \phivec_m \alpha_m - \phivec_m \lamvec_m]\mathbf{u}_m^T \nonumber \\
          & = & D_i - [\phivec_m - \hat{\phivec}_m]\alpha_m \mathbf{u}_m^T \nonumber \\
          & = & D_i - \mathbf{r}_m \alpha_m \mathbf{u}_m^T. \nonumber
\end{eqnarray}

\subsection{Equation \eqref{eqPhi_h}} \label{appendix:pPhi}
\begin{eqnarray}
  \hat{\Psi} = \hat{U} K_i \hat{U}^T
           & = & [U_i - U_{\downarrow m} + \mathbf{u}_m \alpha_m \mathbf{v}_m^T] K_i
              [U_i^T - U_{\downarrow m}^T + \mathbf{v}_m \alpha_m \mathbf{u}_m^T] \nonumber \\
           & = & \, U_i K_i U_i^T - U_i K_i U_{\downarrow m}^T + U_i K_i \mathbf{v}_m \alpha_m \mathbf{u}_m^T
              - U_{\downarrow m} K_i U_i^T \nonumber \\
           &   & + \, U_{\downarrow m} K_i U_{\downarrow m}^T
              - U_{\downarrow m} K_i \mathbf{v}_m \alpha_m \mathbf{u}_m^T
              + \mathbf{u}_m \alpha_m \mathbf{v}_m^T K_i U_i^T \nonumber \\
           &   & - \, \mathbf{u}_m \alpha_m \mathbf{v}_m^T K_i U_{\downarrow m}^T
              + \mathbf{u}_m \alpha_m \mathbf{v}_m^T K_i \mathbf{v}_m \alpha_m \mathbf{u}_m^T \nonumber \\
           & = & \Psi_i - U_i \mathbf{k}_m \lamvec_m \mathbf{u}_m^T + U_i K_i \mathbf{v}_m \alpha_m \mathbf{u}_m^T
                 - \mathbf{u}_m \lamvec_m \mathbf{k}_m^T U_i^T \nonumber \\
           &   & + \, \mathbf{u}_m \lamvec_m K_i(m,m) \lamvec_m \mathbf{u}_m^T
              - \mathbf{u}_m \lamvec_m \mathbf{k}_m^T \mathbf{v}_m \alpha_m \mathbf{u}_m^T \nonumber \\
           &   & + \, \mathbf{u}_m \alpha_m \mathbf{v}_m^T K_i U_i^T - \mathbf{u}_m \alpha_m \mathbf{v}_m^T \mathbf{k}_m \lamvec_m \mathbf{u}_m^T \nonumber \\
           &   & + \, \mathbf{u}_m \alpha_m \mathbf{v}_m^T K_i \mathbf{v}_m \alpha_m \mathbf{u}_m^T \nonumber \\
\hat{\Psi} & = & \Psi_i - (\mathbf{u}_m \hat{\mathbf{u}}_m^T + \hat{\mathbf{u}}_m \mathbf{u}_m^T) \nonumber \\
           &   & + \, \mathbf{u}_m [\lamvec_m \sigma_m^2 \lamvec_m - \lamvec_m \mathbf{k}_m^T \mathbf{v}_m \alpha_m - \alpha_m \mathbf{v}_m^T \mathbf{k}_m \lamvec_m \nonumber \\
           &   & + \, \alpha_m \mathbf{v}_m^T K_i \mathbf{v}_m \alpha_m] \mathbf{u}_m^T, \nonumber
\end{eqnarray}

\bibliographystyle{ieeetr}
\bibliography{BibSource}
\end{document}